\documentclass{article}

\usepackage[preprint]{corl_2026} 

\usepackage{amssymb}
\usepackage{graphicx}
\usepackage{makecell}
\usepackage{multirow}
\usepackage{amsmath}
\usepackage{booktabs}
\usepackage{array}
\usepackage{enumitem}
\usepackage{hyperref}
\usepackage{xcolor}
\usepackage{colortbl}

\usepackage{float}

\usepackage[font=small,skip=3pt]{caption}

\title{H-VLA: Hierarchical Vision-Language-Action Model with Key-Action Reasoning and Motion Planning in a Unified Action Space}

\author{
\textbf{
Xiongfeng Peng$^{1}$,
Lu Xu$^{1}$,
Yandong Wang$^{1}$,
Jiaqian Yu$^{1}$,
Zirui Zheng$^{1}$,} \\
\textbf{
Yamin Mao$^{1}$,
Weiming Li$^{1}$,
Inseop Chung$^{2}$,
Hyun-woong Cho$^{2}$,} \\
\textbf{
Jaewook Yoo$^{2}$,
Dongwook Lee$^{2}$,
Daehyun Ji$^{2}$,
Chao Zhang$^{1}$}\\[0.3em]
$^{1}$Advanced Research Lab, Samsung R\&D Institute China-Beijing (SRCB), China\\
$^{2}$Samsung AI Center, DS Division, South Korea
}

\begin{document}
\maketitle
\vspace{-2.0em}

\begin{figure}[H]
    \centering
    \includegraphics[width=\textwidth]{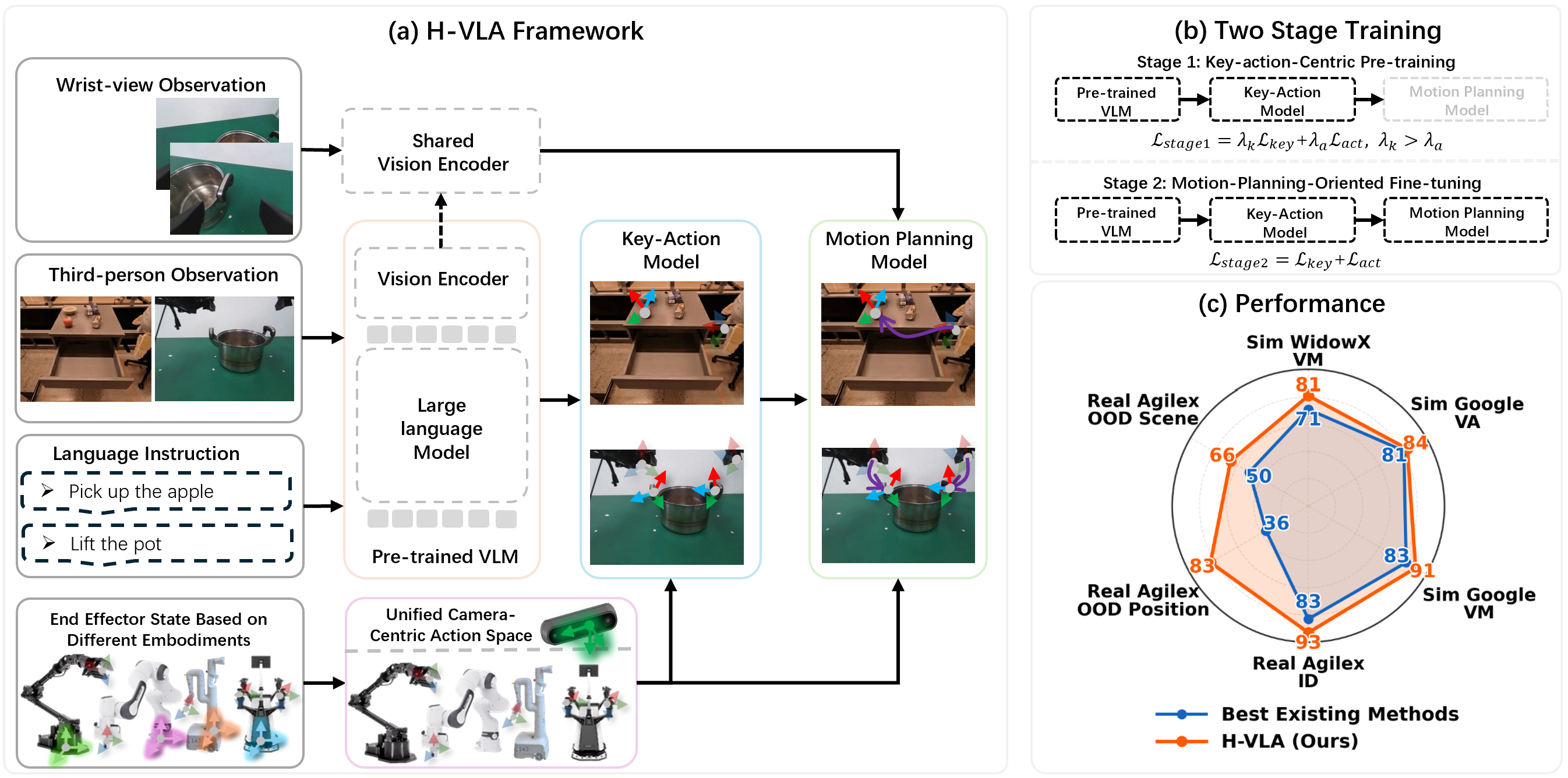}
    \caption{
    \textbf{H-VLA: A hierarchical vision-language-action framework for cross-embodiment robot manipulation.}
    (a) Given third-person RGB observations, optional wrist-view observations, language instructions, and current end-effector states, H-VLA first predicts a structured manipulation \emph{key-action} through a Key-Action Model and then generates future frame-by-frame actions through a Motion Planning Model. A unified camera-centric action space standardizes representations across heterogeneous embodiments and datasets. (b) H-VLA employs a two-stage training strategy, including Key-Action-Centric Pre-training and Motion-Planning-Oriented Fine-tuning. (c) H-VLA achieves strong performance across simulation and real-robot benchmarks.
    }
    \label{fig:overview}
    \vspace{-0.4em}
\end{figure}

\begin{abstract}
Vision-Language-Action (VLA) models have shown strong potential for robotic manipulation, but many existing methods still rely on direct mappings from language and visual observations to dense actions. This formulation can weaken the semantic reasoning capability inherited from pre-trained Vision-Language Models (VLMs), which are mainly optimized for visual-linguistic understanding rather than low-level control, and becomes fragile under spatial variations, including changes in object positions, scene layouts, robot embodiments, and camera viewpoints. To address these limitations, we propose \textbf{H-VLA}, a hierarchical VLA framework that decouples high-level key-action reasoning from low-level motion generation. H-VLA combines a \textbf{Key-Action Model} for predicting a key-action as the next manipulation subgoal, a \textbf{Motion Planning Model} for generating dense future actions conditioned on the predicted key-action, and a \textbf{Unified Camera-Centric Action Space} for consistent representation across datasets, embodiments, and viewpoints. We further adopt a two-stage training strategy that emphasizes key-action reasoning during pre-training and dense motion generation during fine-tuning. Experiments show that H-VLA achieves strong performance on SimplerEnv, reaching 91\% on Google Robot visual matching, 84\% on Google Robot variant aggregation, and 81\% on WidowX visual matching. On Agilex real-robot tasks, H-VLA improves over the strongest baseline by 10, 47, and 16 percentage points under in-distribution, out-of-distribution position, and out-of-distribution scene/object settings, respectively.
\end{abstract}

\keywords{Vision-Language-Action, Robot Manipulation, Hierarchical Control}

\section{Introduction}

Vision-Language-Action (VLA) models have recently emerged as a promising paradigm for robot manipulation, extending large-scale pre-trained Vision-Language Models (VLMs) from perception and reasoning to embodied control~\cite{zitkovich2023rt,kim2024openvla,black2024pi_0,li2024cogact}. By leveraging rich visual semantics and language understanding, VLA models aim to generalize across tasks, objects, and environments more effectively than conventional task-specific policies.

Recent VLA methods have explored several directions toward general robot control. End-to-end methods such as RT-2~\cite{zitkovich2023rt} and OpenVLA~\cite{kim2024openvla} demonstrate that large-scale pre-trained VLMs can transfer visual-language knowledge to action prediction. Continuous-action VLA policies such as CogACT~\cite{li2024cogact} and the $\pi$ series~\cite{black2024pi_0,intelligence2025pi_5,intelligence2025pi_6} improve action quality through diffusion- or flow-based action generation conditioned on VLM representations. Hierarchical or intermediate-reasoning methods, including RT-H~\cite{belkhale2024rt}, RT-Affordance~\cite{nasiriany2025rt}, and ECoT~\cite{zawalski2024robotic}, introduce reasoning traces, affordances, or subgoals to better connect semantic understanding with manipulation. Other work, such as OC-VLA~\cite{zhang2026grounding}, improves robustness by grounding actions in camera coordinates.

Despite recent progress, existing methods still struggle to connect high-level semantic reasoning with low-level motion generation across heterogeneous datasets and embodiments. Many VLA models directly map language and RGB observations to dense actions. Since pre-trained VLMs are optimized for visual and linguistic understanding rather than low-level robot control, direct dense-action prediction can be challenging under changes in object positions, scene layouts, robot embodiments, and camera viewpoints. Although some methods with intermediate reasoning or camera-grounded actions alleviate part of the problem, existing approaches still do not fully resolve the gap between semantic reasoning, motion generation, and cross-embodiment consistency.

To address these limitations, we propose \textbf{H-VLA}, shown in Fig.~\ref{fig:overview}, a hierarchical Vision-Language-Action framework that explicitly separates high-level reasoning from low-level motion generation. Given visual observations, a language instruction, and the current end-effector state, H-VLA first predicts a structured intermediate target, called a \emph{key-action}, which represents the next manipulation subgoal as an end-effector pose and gripper target. A dedicated Motion Planning Model then generates dense future frame-by-frame actions conditioned on this key-action and the current end-effector state. In parallel, H-VLA reformulates current end-effector states, key-actions, and actions in a unified camera-centric action space using calibrated third-person camera extrinsics. This representation standardizes heterogeneous datasets and embodiments, allowing key-action and action learning to benefit from shared supervision across data sources. Since the action space is aligned with the third-person camera frame, it also better matches the visual coordinate system of pre-trained VLMs and helps exploit their visual reasoning capability. We train all model components jointly with a two-stage strategy that emphasizes key-action reasoning during pre-training and dense motion generation during fine-tuning.

This design separates what the robot should achieve from how it should execute the behavior. Its central contribution is the joint formulation of structured key-action reasoning, conditioned dense control, and cross-dataset camera-centric supervision within a two-stage training framework.

Our main contributions are summarized as follows:
\begin{itemize}[leftmargin=1.5em]
    \item We propose a hierarchical VLA framework together with a two-stage training strategy that explicitly decouples high-level key-action reasoning from low-level motion generation.
    \item We introduce a Key-Action Model, a Motion Planning Model, and a Unified Camera-Centric Action Space to bridge semantic reasoning, dense control, and cross-dataset consistency.
    \item We demonstrate strong results on SimplerEnv and a dual-arm real-robot benchmark, achieving 91\% / 84\% / 81\% on Google Robot visual matching (VM) / Google Robot variant aggregation (VA) / WidowX VM, and improving over the strongest Agilex baseline by 10, 47, and 16 points under in-distribution (ID), out-of-distribution (OOD) position, and OOD scene/object settings.
\end{itemize}

\section{Related Work}
\label{sec:related}

\textbf{End-to-End VLAs.}
Recent Vision-Language-Action models extend large-scale pre-trained VLMs to robotic control by mapping visual observations and language instructions to actions. Representative methods such as RT-2~\cite{zitkovich2023rt} and OpenVLA~\cite{kim2024openvla} show that large-scale visual-language pre-training can provide strong generalization for manipulation. Subsequent methods further improve this paradigm through continuous action modeling, larger robot datasets, and richer input interfaces. The $\pi$ series~\cite{black2024pi_0,intelligence2025pi_5,intelligence2025pi_6}, CogACT~\cite{li2024cogact}, TinyVLA~\cite{wen2025tinyvla}, and RDT-1B~\cite{liu2025rdt} enhance action quality, temporal coherence, and precision through flow- or diffusion-style policy modeling. Other generalist systems, including InstructVLA~\cite{yang2025instructvla}, OneTwoVLA~\cite{lin2025onetwovla}, InternVLA-A1~\cite{cai2026internvla}, Interleave-VLA~\cite{fan2025interleave}, UniVLA~\cite{bu2025univla}, CronusVLA~\cite{li2025cronusvla}, GR-3~\cite{cheang2025gr}, Xiaomi-Robotics-0~\cite{cai2026xiaomi}, and ABot-M0~\cite{yang2026abot}, broaden the scope of generalist VLA learning. Despite these advances, most methods still directly map language and RGB observations to dense actions. H-VLA instead introduces an explicit key-action representation to separate semantic reasoning from dense motion generation.

\textbf{Hierarchically Decoupled and Intermediate-Reasoning VLAs.}
Another line of research improves robotic generalization by introducing explicit reasoning, affordance guidance, visual subgoals, or hierarchical decomposition. RT-H~\cite{belkhale2024rt}, RT-Affordance~\cite{nasiriany2025rt}, and ECoT~\cite{zawalski2024robotic} incorporate language-based reasoning traces, affordance predictions, or embodied chain-of-thought reasoning before action prediction. Related work further explores visual subgoal reasoning in CoT-VLA~\cite{zhao2025cot}, sequential affordance reasoning in CoA-VLA~\cite{li2025coa}, image-space paths in HAMSTER~\cite{li2025hamster}, and world-model-generated visual subgoals in VISTA~\cite{long2026scaling}. Other hierarchical or structured VLAs, such as PALM~\cite{liu2026palm}, HiVLA~\cite{yang2026hivla}, ThinkAct~\cite{huang2026thinkact}, and ManualVLA~\cite{gu2025manualvla}, further improve grounding, controllability, and decomposition. ReKep~\cite{huang2024rekep} plans with relational keypoint constraints, while MolmoAct~\cite{lee2025molmoact} conditions actions on depth tokens and image-space waypoints. In contrast to methods that mainly use intermediate language, image-space affordances, or visual subgoals, H-VLA predicts a structured camera-frame end-effector key-action consisting of a 6-DoF pose and gripper state and uses it to condition downstream dense motion planning.

\textbf{Generalist, Steerable, and Camera-Grounded VLA Systems.}
Recent VLA research also explores stronger generalization, efficiency, and deployability through dual-system design, steerability, memory, and camera-grounded actions. RoboDual~\cite{bu2024towards}, OpenHelix~\cite{cui2025openhelix}, Hume~\cite{song2025hume}, and Fast-in-Slow~\cite{chen2025fastinslow} study decompositions between reasoning and execution, while GR00T N1~\cite{bjorck2025gr00t}, MemoryVLA~\cite{shi2025memoryvla}, and LingBot-VLA~\cite{wu2026pragmatic} emphasize scalable generalist control, long-horizon memory, and real-world deployment. OC-VLA~\cite{zhang2026grounding} improves robustness to viewpoint variation by grounding actions in camera coordinates, and $\pi_{0.7}$~\cite{intelligence2026pi_7} advances steerable robotic foundation models through richer context conditioning. These works are complementary to H-VLA. Rather than focusing only on scale, memory, or system-level composition, H-VLA jointly models structured key-action reasoning, camera-centric action representation, and conditioned low-level motion planning.

\begin{figure*}[t]
\centering
\includegraphics[width=0.8\linewidth]{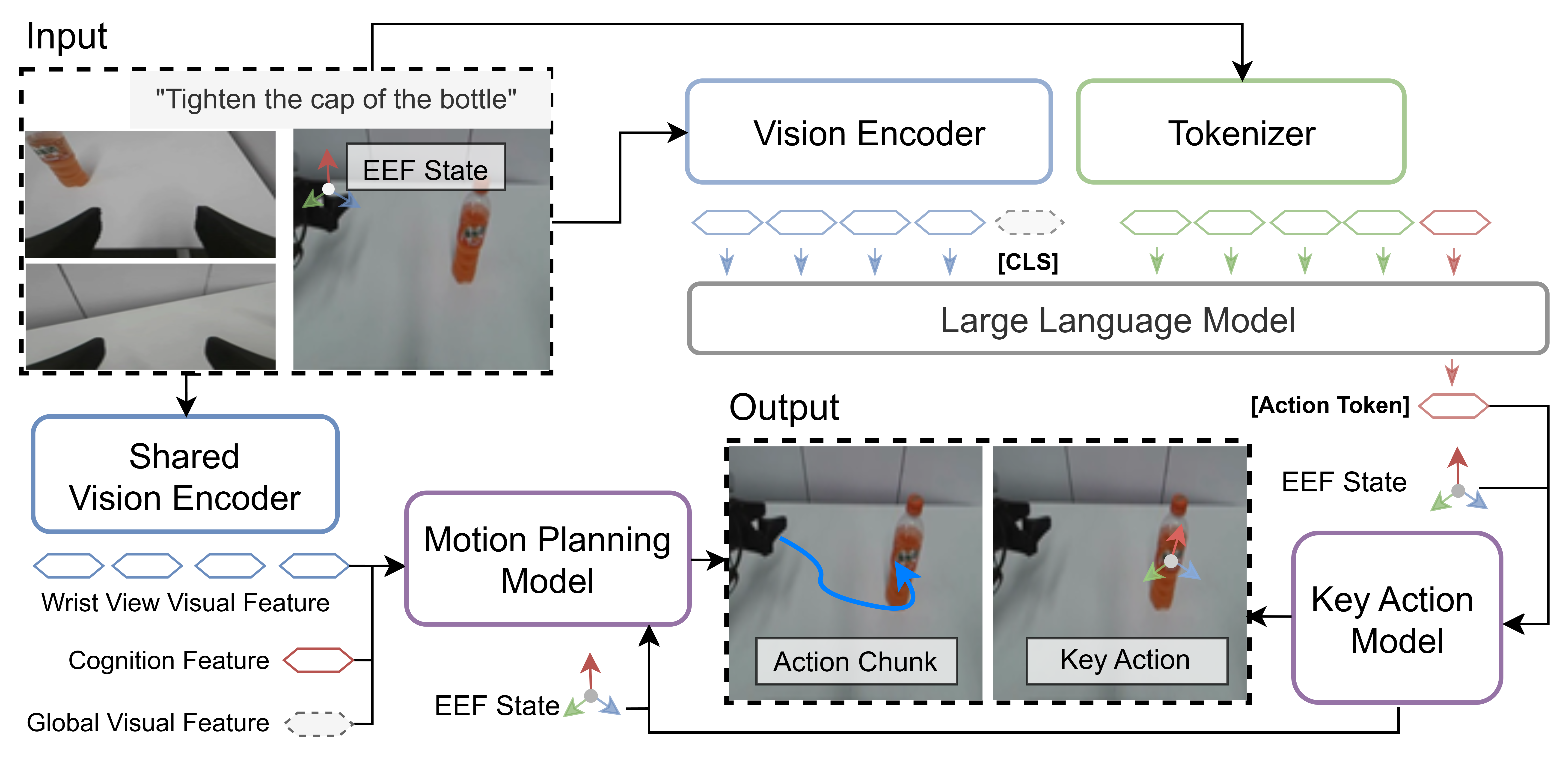}
\caption{
Detailed architecture of H-VLA using the dual-arm task \textit{``tighten the cap of the bottle''} as an example. Given third-person RGB observations, wrist-view observations, language instructions, and current end-effector (EEF) states, a pre-trained VLM extracts the cognition feature from a learnable action token and global visual features from the class (CLS) token. Wrist-view observations are encoded by the vision encoder shared with the VLM, and the availability of left and right wrist cameras is indicated by a wrist-camera mask. The Key-Action Model predicts a structured camera-frame end-effector key-action representing the next manipulation subgoal, while the Motion Planning Model generates dense future action chunks in the same camera-centric action space conditioned on the predicted key-action, cognition feature, global visual feature, wrist-view visual features, and current end-effector state.
}
\label{fig:framework}
\end{figure*}

\section{Method}
\label{sec:method}

\subsection{Overview}
H-VLA is a hierarchical Vision-Language-Action framework that explicitly decouples high-level key-action reasoning from low-level motion generation. As shown in Fig.~\ref{fig:framework}, given third-person RGB observations, wrist-view observations, a language instruction, and the current end-effector state, H-VLA extracts multimodal features with a pre-trained Prismatic-7B Vision-Language Model~\cite{karamcheti2024prismatic}. It then predicts a structured key-action and generates dense future action chunks conditioned on the predicted key-action, multimodal features, and the current end-effector state. Current end-effector states, key-actions, and actions are all represented in a unified camera-centric action space to reduce inconsistencies across heterogeneous datasets, robot embodiments, and viewpoints.

Specifically, third-person RGB observations are encoded by DINOv2~\cite{oquab2023dinov2} and SigLIP~\cite{zhai2023sigmoid}, while the language instruction is tokenized and processed together with visual tokens by the LLaMA-2 language backbone~\cite{touvron2023llama2}. We append a learnable action token to the VLM input sequence and use its final-layer hidden state as the cognition feature. The vision encoder also provides global visual features from the DINOv2 class (CLS) token. Wrist-view observations are encoded by the same vision encoder shared with the VLM. The presence of left and right wrist cameras is specified by a wrist-camera mask, allowing the model to handle missing or unavailable wrist views while using available wrist-view features as local visual cues for motion planning. Both the Key-Action Model and the Motion Planning Model are implemented as diffusion-based transformer modules~\cite{peebles2023scalable}.

\subsection{Problem Formulation}
At each robot timestep $t$, H-VLA takes as input third-person RGB observations $\mathbf{o}^{3rd}_t$, optional wrist-view observations $\mathbf{o}^{wrist}_t$, a language instruction $\mathbf{l}$, and the current end-effector state $\mathbf{s}_t$. Current end-effector states and key-actions are represented as absolute Cartesian end-effector position, Euler orientation, and gripper state, while action commands use the same dimensional layout but encode frame-to-frame position/orientation deltas with an absolute gripper state. For a single-arm robot, the state/key-action pose vector is represented as $\mathbf{u}_t=[x,y,z,\phi,\theta,\psi,g]\in\mathbb{R}^{7}$, where $(x,y,z)$ denotes Cartesian position, $(\phi,\theta,\psi)$ denotes Euler orientation, and $g$ denotes the gripper state. For dual-arm robots, the representation is extended by concatenating the left- and right-arm vectors, i.e., $\mathbf{u}_t=[\mathbf{u}^{L}_t,\mathbf{u}^{R}_t]\in\mathbb{R}^{14}$. The 7-dimensional representation defines the physical single-arm interface for states, key-actions, and action commands, while H-VLA adopts a shared 14-dimensional model interface for mixed single-arm and dual-arm training.
Instead of directly mapping observations and language to dense actions, H-VLA factorizes policy prediction into two conditional generation stages. The Key-Action Model models the conditional distribution $p_{\theta}^{k}(\mathbf{k}_t \mid \mathbf{c}^{k}_t)$, where $\mathbf{c}^{k}_t$ is defined in Sec.~\ref{sec:key_action}. The Motion Planning Model models the conditional distribution $p_{\theta}^{a}(\mathbf{A}_t \mid \mathbf{c}^{a}_t)$, where $\mathbf{c}^{a}_t$ is defined in Sec.~\ref{sec:motion_planning}. Here, $\mathbf{A}_t =[\Delta\mathbf{a}_{t+1},\Delta\mathbf{a}_{t+2},\dots,\Delta\mathbf{a}_{t+H}]$, and $H$ denotes the action horizon.

\subsection{Key-Action Model}
\label{sec:key_action}

The Key-Action Model predicts the next manipulation subgoal as a structured target: the absolute end-effector pose and gripper state at the next salient manipulation event in the unified camera-centric action space. Let $\mathbf{f}^{cog}_t$ denote the cognition feature extracted from the learnable VLM action token. The condition for the Key-Action Model is constructed as
\begin{equation}
\mathbf{c}^{k}_t =
\operatorname{concat}
\left(
\mathbf{f}^{cog}_t,
\phi_s(\mathbf{s}_t)
\right),
\end{equation}
where $\phi_s(\cdot)$ is a two-layer MLP that encodes the current end-effector state.

We formulate key-action prediction as a standard diffusion denoising problem~\cite{ho2020denoising, peebles2023scalable}. Let $\mathbf{k}_t$ denote the ground-truth key-action associated with timestep $t$ in the unified camera-centric action space. During training, Gaussian noise is added to this clean target to obtain $\mathbf{k}_{t,i}$ at diffusion step $i$, and the model is optimized to predict the injected noise:
\begin{equation}
\mathcal{L}_{\text{key}} =
\mathbb{E}_{i,\boldsymbol{\epsilon}^{k}}
\left[
\left\|
\epsilon^{k}_{\theta}
(\mathbf{k}_{t,i}, i, \mathbf{c}^{k}_t)
-
\boldsymbol{\epsilon}^{k}
\right\|_2^2
\right].
\end{equation}

Ground-truth key-actions are constructed from gripper-state changes in demonstration trajectories. Specifically, the frame immediately following each gripper-state transition (opening or closing) is defined as a key-action. In addition, the final frame of each demonstration trajectory is always treated as a key-action to represent the terminal manipulation state. For each timestep, the target key-action is the closest future key-action defined by either a gripper-state transition or the trajectory end. For dual-arm trajectories, key-actions are extracted independently for the left and right arms and represented as absolute end-effector poses in the unified camera-centric action space. These event labels provide an automatic approximation to the next manipulation subgoal.

\subsection{Motion Planning Model}
\label{sec:motion_planning}

The Motion Planning Model generates dense future action chunks conditioned on a key-action, current end-effector state, and multimodal features. A key-action specifies an intermediate target without fully determining the intervening trajectory; the learned motion model combines this target with task and scene information to generate context-dependent execution. Its condition follows the information flow in Fig.~\ref{fig:framework}: the cognition feature $\mathbf{f}^{cog}_t$ from the learnable VLM action token, the global visual feature $\mathbf{f}^{cls}_t$ from the CLS token, wrist-view visual features encoded by the vision encoder shared with the VLM, and the encoded current-state--key-action pair are concatenated:
\begin{equation}
\mathbf{c}^{a}_t =
\operatorname{concat}
\left(
\mathbf{f}^{cog}_t,
\mathbf{f}^{cls}_t,
\phi_w(\mathbf{o}^{wrist}_{t}),
\phi_m(\mathbf{m}^{wrist}_t),
\phi_{ak}\left([\mathbf{s}_t, \mathbf{k}^{\mathrm{cond}}_t]\right)
\right),
\end{equation}
where $\phi_w(\cdot)$ encodes left and right wrist-view observations using the shared vision encoder, and $\phi_{ak}(\cdot)$ is a two-layer MLP that encodes the current end-effector state together with the key-action condition. During training, $\mathbf{k}^{\mathrm{cond}}_t$ is the ground-truth key-action, while at inference it is the key-action predicted by the Key-Action Model. $\mathbf{m}^{wrist}_t\in\{0,1\}^{2}$ indicates left/right wrist-camera availability, and $\phi_m(\cdot)$ is a two-layer MLP that encodes this mask and appends it to the motion-planning condition. When a wrist view is unavailable, its visual feature is zero-filled and masked out while preserving a fixed model interface.

The future action chunk associated with timestep $t$ is defined as
\begin{equation}
\mathbf{A}_t=
[\Delta\mathbf{a}_{t+1},
\Delta\mathbf{a}_{t+2},
\dots,
\Delta\mathbf{a}_{t+H}].
\end{equation}
The horizon $H$ is fixed independently of the distance to the predicted key-action. Longer motions are realized through additional closed-loop policy queries. For each action command, Cartesian position and Euler orientation are represented as frame-to-frame deltas between consecutive end-effector poses expressed in the camera frame, while the gripper remains an absolute state. The Motion Planning Model is optimized using the same diffusion denoising objective on action chunks:
\begin{equation}
\mathcal{L}_{\text{act}} =
\mathbb{E}_{i,\boldsymbol{\epsilon}^{a}}
\left[
\left\|
\epsilon^{a}_{\theta}
(\mathbf{A}_{t,i}, i, \mathbf{c}^{a}_t)
-
\boldsymbol{\epsilon}^{a}
\right\|_2^2
\right].
\end{equation}

\subsection{Unified Camera-Centric Action Space}

A key component of H-VLA is the unified camera-centric action space. Although different datasets and robot platforms define end-effector poses in different robot-base frames, H-VLA reformulates current end-effector states, key-actions, and future action chunks in the unified third-person camera frame. Specifically, given the base-to-camera transformation $\mathbf{T}^{c}_{b}$ and the end-effector pose in the robot base frame $\mathbf{T}^{b}_{ee}$, the corresponding camera-frame pose is computed as
\begin{equation}
\mathbf{T}^{c}_{ee}=
\mathbf{T}^{c}_{b}\mathbf{T}^{b}_{ee}.
\end{equation}

The Cartesian position and Euler orientation are extracted from $\mathbf{T}^{c}_{ee}$ to define current end-effector states and key-actions, while the position and orientation components of future actions are represented as frame-to-frame deltas between consecutive end-effector poses expressed in the camera frame, with the gripper represented as an absolute state. This transformation is applied to all labels and states across datasets, aligning robot control labels with the visual observation frame and improving cross-dataset consistency across embodiments and viewpoints.

\subsection{Two-Stage Training Strategy}

To balance generalizable reasoning and downstream motion adaptation, H-VLA adopts a two-stage training strategy, as illustrated in Fig.~\ref{fig:overview}. The first stage, \textbf{Key-Action-Centric Pre-training}, emphasizes transferable key-action reasoning on heterogeneous manipulation data. The second stage, \textbf{Motion-Planning-Oriented Fine-tuning}, adapts the model to downstream tasks with emphasis on dense motion generation. In both stages, all model parameters are optimized jointly, but the relative emphasis of the losses changes.

\paragraph{Stage 1: Key-Action-Centric Pre-training.}

In the first stage, training emphasizes transferable key-action reasoning on heterogeneous manipulation data:
\begin{equation}
\mathcal{L}_{\text{stage1}} =
\lambda_k \mathcal{L}_{\text{key}}
+
\lambda_a \mathcal{L}_{\text{act}},
\quad
\lambda_k > \lambda_a.
\end{equation}

This stage encourages the model to learn robust object-centric reasoning and the next manipulation subgoal representation.

\paragraph{Stage 2: Motion-Planning-Oriented Fine-tuning.}

In the second stage, the model is adapted to downstream manipulation data while retaining both key-action and dense-action supervision:
\begin{equation}
\mathcal{L}_{\text{stage2}} =
\mathcal{L}_{\text{key}}
+
\mathcal{L}_{\text{act}}.
\end{equation}

Equal weighting increases the relative emphasis on dense-action learning compared with Stage~1 while preserving key-action supervision during downstream adaptation.

\section{Experiments}
\label{sec:experiments}

We organize the experiments into four parts. First, we describe the pre-training and fine-tuning settings used to adapt H-VLA from heterogeneous single-arm and dual-arm datasets to downstream benchmarks. Second, we evaluate simulation transfer on SimplerEnv~\cite{li2024evaluating}. Third, we evaluate real-robot adaptation on an Agilex dual-arm platform. Finally, we conduct ablation studies on key-action reasoning and motion planning, the camera-centric action space, and the proposed training strategy.

\subsection{Pre-training and Fine-tuning}

H-VLA is first pre-trained on a mixed set of single-arm and dual-arm datasets, comprising approximately 183K robot trajectories. The dual-arm data include RoboCOIN Cobot Magic~\cite{wu2025robocoin}, while the single-arm data include DROID~\cite{khazatsky2024droid}, FractalData~\cite{brohan2022rt}, and BridgeDataV2~\cite{walke2023bridgedata}. Mixed pre-training exposes H-VLA to diverse embodiments, task semantics, camera viewpoints, and action distributions within the unified camera-centric action space. The mixed pre-training stage is conducted using 8 NVIDIA H100 GPUs for approximately three days and one epoch. Unless otherwise specified, pre-training uses a constant learning rate of $2\times10^{-5}$ and a batch size of 256. During Stage~1 Key-Action-Centric Pre-training, we set $\lambda_k=10$ and $\lambda_a=1$ to emphasize transferable key-action reasoning over dense motion generation. All parameters are updated during both pre-training and fine-tuning.

Starting from the mixed pre-trained model, we conduct two groups of downstream fine-tuning. For simulation transfer, we separately fine-tune H-VLA on FractalData and BridgeDataV2, and evaluate the resulting policies on SimplerEnv~\cite{li2024evaluating}. FractalData fine-tuning is performed on 8 NVIDIA H100 GPUs for approximately three days and four epochs. BridgeDataV2 fine-tuning is performed on 8 NVIDIA H100 GPUs for approximately three days and six epochs.

For real-robot adaptation, we collect 150 demonstrations on an Agilex dual-arm robot across three tasks: \emph{PickBottles}, \emph{HandOver}, and \emph{LiftPot}, with 50 demonstrations per task. Starting from the same mixed pre-trained model, the Agilex model is fine-tuned using 4 NVIDIA H100 GPUs for approximately one day and 100 epochs. This corresponds to Stage~2 Motion-Planning-Oriented Fine-tuning, which adapts the model to downstream tasks while preserving key-action supervision.

\subsection{Simulation Evaluation on SimplerEnv}

We first evaluate whether mixed pre-training transfers effectively to standard simulated manipulation benchmarks. We report results under the SimplerEnv benchmark~\cite{li2024evaluating}. For the Google Robot benchmark, we report both Visual Matching (VM) and Variant Aggregation (VA); for the WidowX benchmark, we report VM results.
We compare H-VLA with representative VLA baselines, including RT-1-X~\cite{o2024open}, RT-2-X~\cite{o2024open}, Octo~\cite{team2024octo}, OpenVLA~\cite{kim2024openvla}, RoboVLM~\cite{li2024towards}, SpatialVLA~\cite{qu2025spatialvla}, CogACT~\cite{li2024cogact}, EMMA-X~\cite{sun2025emma}, GR00T N1.5~\cite{bjorck2025gr00t}, MolmoAct~\cite{lee2025molmoact}, $\pi_0$~\cite{black2024pi_0}, DAM-VLA~\cite{peng2026dam}, and X-VLA~\cite{zheng2025x}. Table~\ref{tab:simplerenv_all} distinguishes results evaluated by us using the SimplerEnv evaluation setup (${}^{\dagger}$) from published results (${}^{\ddagger}$) and records the corresponding training settings.

\begin{table*}[t]
\setlength{\abovecaptionskip}{0.15cm}
\setlength{\belowcaptionskip}{-0.1cm}
\centering
\setlength{\tabcolsep}{3pt}
\resizebox{1.0\textwidth}{!}{
\begin{tabular}{l|ccccc|ccccc|ccccc}
\toprule
& \multicolumn{5}{c|}{Google Robot (VM)} & \multicolumn{5}{c|}{Google Robot (VA)} & \multicolumn{5}{c}{WidowX (VM)} \\
Policy & \makecell{Coke\\Can} & \makecell{Move\\Near} & \makecell{Open/\\Close} & \makecell{Drawer+\\Apple} & Avg & \makecell{Coke\\Can} & \makecell{Move\\Near} & \makecell{Open/\\Close} & \makecell{Drawer+\\Apple} & Avg & \makecell{Spoon\\Towel} & \makecell{Carrot\\Plate} & Stack & \makecell{Eggplant\\Basket} & Avg \\
\midrule
${}^{\dagger}$\,RT-1-X~\cite{o2024open} (PT) & 59\% & 33\% & 56\% & 17\% & 41\% & 49\% & 33\% & 29\% & 10\% & 30\% & 4\% & 8\% & 0\% & 0\% & 3\% \\
${}^{\ddagger}$\,RT-2-X~\cite{o2024open} (PT) & 79\% & 78\% & 25\% & 4\% & 46\% & 82\% & 79\% & 35\% & 21\% & 54\% & -- & -- & -- & -- & -- \\
${}^{\dagger}$\,Octo~\cite{team2024octo} (PT) & 18\% & 4\% & 24\% & 0\% & 11\% & 1\% & 4\% & 1\% & 0\% & 1\% & 4\% & 8\% & 0\% & 38\% & 13\% \\
${}^{\dagger}$\,OpenVLA~\cite{kim2024openvla} (PT) & 14\% & 51\% & 48\% & 0\% & 28\% & 64\% & 64\% & 19\% & 1\% & 37\% & 4\% & 0\% & 0\% & 4\% & 2\% \\
${}^{\ddagger}$EMMA-X~\cite{sun2025emma} (PT) & 2\% & 3\% & 18\% & -- & -- & 5\% & 7\% & 21\% & -- & -- & -- & -- & -- & -- & -- \\
${}^{\dagger}$\,CogACT~\cite{li2024cogact} (PT) & 92\% & 82\% & 75\% & 39\% & 72\% & 96\% & 84\% & 29\% & 40\% & 62\% & 63\% & 50\% & 25\% & 71\% & 52\% \\
\midrule
${}^{\ddagger}$\,RoboVLM~\cite{li2024towards} (FT) & 77\% & 62\% & 44\% & -- & -- & 76\% & 60\% & 11\% & -- & -- & 29\% & 25\% & 13\% & 58\% & 31\% \\
${}^{\ddagger}$\,SpatialVLA~\cite{qu2025spatialvla} (FT) & 86\% & 78\% & 57\% & -- & -- & 88\% & 73\% & 42\% & -- & -- & 17\% & 25\% & 29\% & \textbf{100\%} & 43\% \\
${}^{\ddagger}$GR00T N1.5~\cite{bjorck2025gr00t} (FT) & 69\% & 69\% & 36\% & -- & -- & 47\% & 63\% & 18\% & -- & -- & -- & -- & -- & -- & -- \\
${}^{\ddagger}$MolmoAct~\cite{lee2025molmoact} (FT) & 78\% & 77\% & 60\% & -- & -- & 76\% & 61\% & \textbf{79\%} & -- & -- & -- & -- & -- & -- & -- \\
${}^{\ddagger}$\,$\pi_0$~\cite{black2024pi_0} (FT) & 73\% & 65\% & 38\% & -- & -- & 75\% & 64\% & 26\% & -- & -- & -- & -- & -- & -- & -- \\
${}^{\ddagger}$\,$\pi_0^{*}$~\cite{black2024pi_0} (T) & 89\% & 81\% & 55\% & 53\% & 70\% & -- & -- & -- & -- & -- & 62\% & 59\% & 24\% & 81\% & 57\% \\
${}^{\dagger}$\,X-VLA~\cite{zheng2025x} (FT) & 96\% & 84\% & 66\% & 32\% & 70\% & 83\% & 70\% & 45\% & 35\% & 58\% & 83\% & 63\% & 42\% & 0\% & 47\% \\
${}^{\dagger}$\,DAM-VLA~\cite{peng2026dam} (T) & 96\% & 84\% & 75\% & 78\% & 83\% & \textbf{98\%} & 74\% & 68\% & \textbf{84\%} & 81\% & 88\% & 71\% & 25\% & \textbf{100\%} & 71\% \\
${}^{\dagger}$\,\textbf{H-VLA (Ours)} (FT) & \textbf{98\%} & \textbf{90\%} & \textbf{80\%} & \textbf{98\%} & \textbf{91\%} & \textbf{98\%} & \textbf{85\%} & \textbf{79\%} & 72\% & \textbf{84\%} & \textbf{96\%} & \textbf{88\%} & \textbf{58\%} & 83\% & \textbf{81\%} \\
\bottomrule
\end{tabular}}
\caption{
Task-wise SimplerEnv success rates. For Google Robot, we report Visual Matching (VM) and Variant Aggregation (VA); for WidowX, we report VM. The best result in each column is shown in bold. Methods with missing task results are marked as -- and are not compared on the corresponding average. Task names are abbreviated. PT: pre-training only; FT: fine-tuning after robot-policy pre-training; T: downstream training without robot-data pre-training. $\pi_0^{*}$ refers to open-pi-zero. ${}^{\dagger}$: evaluated by us; ${}^{\ddagger}$: reported in prior work.
}
\label{tab:simplerenv_all}
\end{table*}

Table~\ref{tab:simplerenv_all} reports the full SimplerEnv results. H-VLA achieves the strongest average performance across all three benchmark splits, reaching \textbf{91\%} on Google Robot VM, \textbf{84\%} on Google Robot VA, and \textbf{81\%} on WidowX VM. Compared with strong baselines, H-VLA particularly improves tasks requiring structured intermediate reasoning followed by multi-step motion execution. For example, \emph{Drawer+Apple} involves two sequential sub-tasks, opening the top drawer and placing the apple, making it more challenging than single-step tasks. H-VLA achieves \textbf{98\%} on this task under Google Robot VM, substantially outperforming DAM-VLA~\cite{peng2026dam} at 78\%. These results suggest that explicit key-action reasoning effectively bridges semantic reasoning and dense action generation, enabling better transfer across different embodiments and manipulation tasks.

\subsection{Real-Robot Evaluation on Agilex}

We next evaluate H-VLA on a dual-arm Agilex robot. The real-robot dataset contains 150 demonstrations across three tasks: \emph{PickBottles}, where both arms grasp one bottle each; \emph{HandOver}, where one arm transfers a bottle to the other arm; and \emph{LiftPot}, where both arms cooperatively lift a pot. Each task contains 50 demonstrations.

For a fair comparison, H-VLA, $\pi_0$~\cite{black2024pi_0}, $\pi_{0.5}$~\cite{intelligence2025pi_5}, CogACT~\cite{li2024cogact}, and MolmoAct2~\cite{fang2026molmoact2} are all fine-tuned on the same 150-demonstration Agilex dataset, with each method producing one unified policy for all three tasks. H-VLA is fine-tuned from the mixed pre-trained model. $\pi_0$ and $\pi_{0.5}$ are initialized from their respective OpenPI base checkpoints, CogACT uses its OXE-pretrained checkpoint, and MolmoAct2 is fine-tuned from the MolmoAct2-BimanualYAM checkpoint. All results are evaluated by us using common task definitions and success criteria.

H-VLA, $\pi_0$, $\pi_{0.5}$, and MolmoAct2 use one third-person view and two wrist views, while our CogACT baseline uses only the third-person view. All five models are evaluated under ID and OOD Position; OOD Scene/Object is additionally evaluated for H-VLA, $\pi_0$, and $\pi_{0.5}$. OOD Position modifies object placement while preserving the task setup. OOD Scene/Object introduces broader variations including background changes and novel target objects. These shifts are defined relative to the Agilex fine-tuning distribution, with OOD test positions excluded from fine-tuning.

\begin{table*}[t]
\setlength{\abovecaptionskip}{0.15cm}
\setlength{\belowcaptionskip}{-0.1cm}
\centering
\setlength{\tabcolsep}{3pt}
\resizebox{1.0\textwidth}{!}{
\begin{tabular}{lcccc|cccc|cccc}
\toprule
& \multicolumn{4}{c|}{ID} & \multicolumn{4}{c|}{OOD Position} & \multicolumn{4}{c}{OOD Scene/Object} \\
\cmidrule(lr){2-5}\cmidrule(lr){6-9}\cmidrule(lr){10-13}
Policy & PickBottles & HandOver & LiftPot & Avg & PickBottles & HandOver & LiftPot & Avg & PickBottles & HandOver & LiftPot & Avg \\
\midrule
CogACT~\cite{li2024cogact} & 47\% & 53\% & 50\% & 50\% & 20\% & 33\% & 20\% & 24\% & -- & -- & -- & -- \\
MolmoAct2~\cite{fang2026molmoact2} & 80\% & 60\% & 70\% & 70\% & 53\% & 13\% & 40\% & 36\% & -- & -- & -- & -- \\
$\pi_0$~\cite{black2024pi_0} & 60\% & \textbf{100\%} & \textbf{90\%} & 83\% & 33\% & 30\% & 30\% & 31\% & 37\% & 23\% & \textbf{72\%} & 44\% \\
$\pi_{0.5}$~\cite{intelligence2025pi_5} & 85\% & 80\% & 80\% & 82\% & 23\% & 37\% & 33\% & 31\% & 47\% & 37\% & 67\% & 50\% \\
\textbf{H-VLA} \textbf{(Ours)} & \textbf{100\%} & 90\% & \textbf{90\%} & \textbf{93\%} & \textbf{77\%} & \textbf{93\%} & \textbf{80\%} & \textbf{83\%} & \textbf{73\%} & \textbf{63\%} & 61\% & \textbf{66\%} \\
\bottomrule
\end{tabular}}
\caption{
Real-robot success rates on the Agilex dual-arm benchmark under in-distribution (ID), out-of-distribution (OOD) position shifts, and out-of-distribution (OOD) scene/object shifts.
}
\label{tab:agilex_real}
\vspace{-0.2em}
\end{table*}

\begin{table*}[t]
\centering
{
\setlength{\tabcolsep}{3pt}
\resizebox{1.0\textwidth}{!}{
\begin{tabular}{lcccc|c|cccccc}
\toprule
& \multicolumn{4}{c|}{Model Size} & \multicolumn{1}{c|}{Pretraining} & \multicolumn{6}{c}{Inference} \\
\cmidrule(lr){2-5}\cmidrule(lr){7-12}
Policy & VLM & KAM & AM/MPM & Total & Robot Data & Precision & VLM & KAM & AM/MPM & Total & Chunk \\
\midrule
$\pi_0$~\cite{black2024pi_0} & 3.0B & -- & 300M & 3.3B & $>1000$K & BF16 (mixed) & 49 ms & -- & 22 ms & 73 ms & 50 \\
$\pi_{0.5}$~\cite{intelligence2025pi_5} & 3.0B & -- & 300M & 3.3B & $>1000$K & BF16 (mixed) & 48 ms & -- & 24 ms & 73 ms & 50 \\
MolmoAct2~\cite{fang2026molmoact2} & 4.9B & -- & 621M & 5.5B & $>300$K & BF16 & 90 ms & -- & 320 ms & 420 ms & 30 \\
CogACT~\cite{li2024cogact} & 7.5B & -- & 89M & 7.6B & $\geq400$K & BF16 & 57 ms & -- & 18 ms & 76 ms & 16 \\
\textbf{H-VLA (Ours)} & 7.5B & 89M & 89M & 7.7B & 183K & BF16 & 54 ms & 17 ms & 17 ms & 103 ms & 16 \\
\bottomrule
\end{tabular}}
\caption{Model size, robot pre-training data, and RTX~4090 inference latency. Robot data are reported in trajectories (K: thousands); $>$ and $\geq$ denote conservative estimates or lower bounds. KAM: Key-Action Model; AM/MPM: Action Model/Motion Planning Model. Total denotes end-to-end latency; for H-VLA, it also includes left/right wrist-RGB encoding that is not included in the VLM/KAM/MPM module times. Chunk is the training action-chunk length and the maximum per-query inference horizon.}
\label{tab:model_size_inference}
}
\end{table*}

Table~\ref{tab:agilex_real} shows that H-VLA performs strongly across all three Agilex tasks and maintains high success rates under both position and scene/object shifts. Under ID, H-VLA achieves the highest average success rate of \textbf{93\%}, outperforming the strongest baseline, $\pi_0$ at 83\%, by 10 percentage points. Under OOD Position, H-VLA maintains an average success rate of \textbf{83\%}, compared with 36\% for MolmoAct2 and 31\% for both $\pi_0$ and $\pi_{0.5}$, yielding a 47-point improvement over the strongest evaluated baseline. Under the more challenging OOD Scene/Object setting, H-VLA achieves the highest average success rate of \textbf{66\%}, exceeding $\pi_{0.5}$ at 50\% and $\pi_0$ at 44\% by 16 and 22 percentage points, respectively. In particular, H-VLA achieves 73\% and 63\% on PickBottles and HandOver under OOD Scene/Object, while its advantage is task-dependent: $\pi_0$ and $\pi_{0.5}$ perform better on LiftPot. These results demonstrate that H-VLA retains strong performance under substantial spatial and visual distribution shifts, especially on PickBottles and HandOver.

Table~\ref{tab:model_size_inference} compares model size, robot pre-training data, and RTX~4090 latency. H-VLA has 7.7B parameters, compared with 3.3B for $\pi_0$/$\pi_{0.5}$, 5.5B for MolmoAct2, and 7.6B for CogACT. $\pi_0$ combines OXE Magic Soup, BridgeDataV2, DROID, and 903M proprietary robot timesteps~\cite{black2024pi_0,team2024octo,walke2023bridgedata,khazatsky2024droid}; $\pi_{0.5}$ further includes large-scale mobile-manipulator, multi-environment, and cross-embodiment robot data~\cite{intelligence2025pi_5,o2024open}. MolmoAct2 uses BimanualYAM, SO100/101, DROID, and additional robot datasets including BC-Z, BridgeDataV2, RT-1, and MolmoAct~\cite{fang2026molmoact2}; CogACT is pre-trained on about 400K OXE trajectories~\cite{li2024cogact}. H-VLA uses 183K trajectories from RoboCOIN, DROID, BridgeDataV2, and FractalData~\cite{wu2025robocoin,khazatsky2024droid,walke2023bridgedata,brohan2022rt}. On RTX~4090, H-VLA takes 103\,ms for a 16-step chunk, versus 73\,ms for $\pi_0$/$\pi_{0.5}$ (50 steps), 76\,ms for CogACT (16 steps), and 420\,ms for MolmoAct2 (30 steps). For H-VLA, the 15\,ms gap between the 88\,ms sum of VLM/KAM/MPM module times and the 103\,ms end-to-end total comes from encoding the left and right wrist RGB views, which is included only in Total. Chunk is the action-chunk length used in training and the maximum number of future actions produced by one inference query.

\subsection{Ablation Study}

We conduct ablation studies on SimplerEnv under the SimplerEnv evaluation setup to examine each major component in H-VLA. Starting from the complete H-VLA, we progressively remove key design components. Specifically, \textbf{H-VLA} denotes the full model. \textbf{w/o MPT+TST} removes mixed single-/dual-arm pre-training and the two-stage training strategy. \textbf{w/o CC} further removes the unified camera-centric action space. Finally, \textbf{w/o KA} removes the Key-Action Model and key-action conditioning while retaining the Motion Planning Model as a direct dense-action predictor. We also evaluate \textbf{H-VLA (w/o Stage-2 FT)}, i.e., the checkpoint after Key-Action-Centric Pre-training.

\begin{table*}[t]
\setlength{\abovecaptionskip}{0.15cm}
\setlength{\belowcaptionskip}{-0.1cm}
\centering
\setlength{\tabcolsep}{3pt}
\resizebox{1.0\textwidth}{!}{
\begin{tabular}{l|ccccc|ccccc|ccccc}
\toprule
& \multicolumn{5}{c|}{Google Robot (VM)} & \multicolumn{5}{c|}{Google Robot (VA)} & \multicolumn{5}{c}{WidowX (VM)} \\
Variant & \makecell{Coke\\Can} & \makecell{Move\\Near} & \makecell{Open/\\Close} & \makecell{Drawer+\\Apple} & Avg & \makecell{Coke\\Can} & \makecell{Move\\Near} & \makecell{Open/\\Close} & \makecell{Drawer+\\Apple} & Avg & \makecell{Spoon\\Towel} & \makecell{Carrot\\Plate} & Stack & \makecell{Eggplant\\Basket} & Avg \\
\midrule
\textbf{H-VLA (w/o Stage-2 FT)} & 75\% & 68\% & 74\% & 73\% & 73\% & 77\% & 36\% & 63\% & 21\% & 49\% & 50\% & 63\% & 13\% & 88\% & 53\% \\
\textbf{H-VLA} & \textbf{98\%} & 90\% & \textbf{80\%} & \textbf{98\%} & \textbf{91\%} & \textbf{98\%} & \textbf{85\%} & 79\% & \textbf{72\%} & \textbf{84\%} & \textbf{96\%} & \textbf{88\%} & \textbf{58\%} & 83\% & \textbf{81\%} \\
\midrule
w/o MPT+TST & \textbf{98\%} & \textbf{91\%} & 72\% & 80\% & 85\% & 94\% & 84\% & \textbf{83\%} & 63\% & 81\% & 71\% & 83\% & 42\% & \textbf{100\%} & 74\% \\
w/o MPT+TST+CC & 91\% & 72\% & 69\% & 38\% & 68\% & 93\% & 70\% & 39\% & 33\% & 59\% & 71\% & 63\% & 42\% & 96\% & 68\% \\
w/o MPT+TST+CC+KA & 95\% & 59\% & 53\% & 18\% & 56\% & 93\% & 68\% & 45\% & 34\% & 60\% & 58\% & 54\% & 13\% & 92\% & 54\% \\
\bottomrule
\end{tabular}}
\caption{
Task-wise ablation results on SimplerEnv~\cite{li2024evaluating} under the SimplerEnv evaluation setup. \textbf{MPT} denotes mixed single-/dual-arm pre-training, \textbf{TST} denotes two-stage training, \textbf{CC} denotes the unified camera-centric action space, and \textbf{KA} denotes the Key-Action Model.
}
\label{tab:ablation_all}
\end{table*}

Table~\ref{tab:ablation_all} shows that, despite emphasizing key-action learning ($\lambda_k=10,\lambda_a=1$), H-VLA after Stage~1 already achieves 73\%, 49\%, and 53\% on Google Robot VM, Google Robot VA, and WidowX VM, respectively, without Stage-2 fine-tuning. This supports joint key-action and dense-action learning, with key-action-centric pre-training producing effective motion predictions. Removing mixed single-/dual-arm pre-training and the two-stage training strategy reduces average success rate from \textbf{91\%} to 85\% on Google Robot VM, from \textbf{84\%} to 81\% on Google Robot VA, and from \textbf{81\%} to 74\% on WidowX VM, showing the importance of mixed pre-training and staged optimization for cross-embodiment transfer. Further removing the camera-centric action space leads to a larger drop, especially on Google Robot VM and VA, where average success rates decrease to 68\% and 59\%. Removing KA reduces Google Robot VM from 68\% to 56\% and WidowX VM from 68\% to 54\%, while Google Robot VA changes from 59\% to 60\%. Thus, key-action conditioning benefits two of the three settings even without mixed pre-training or camera-centric actions.

\subsection{Qualitative Results}

\begin{figure*}[t]
    \centering
    \includegraphics[width=1.0\textwidth]{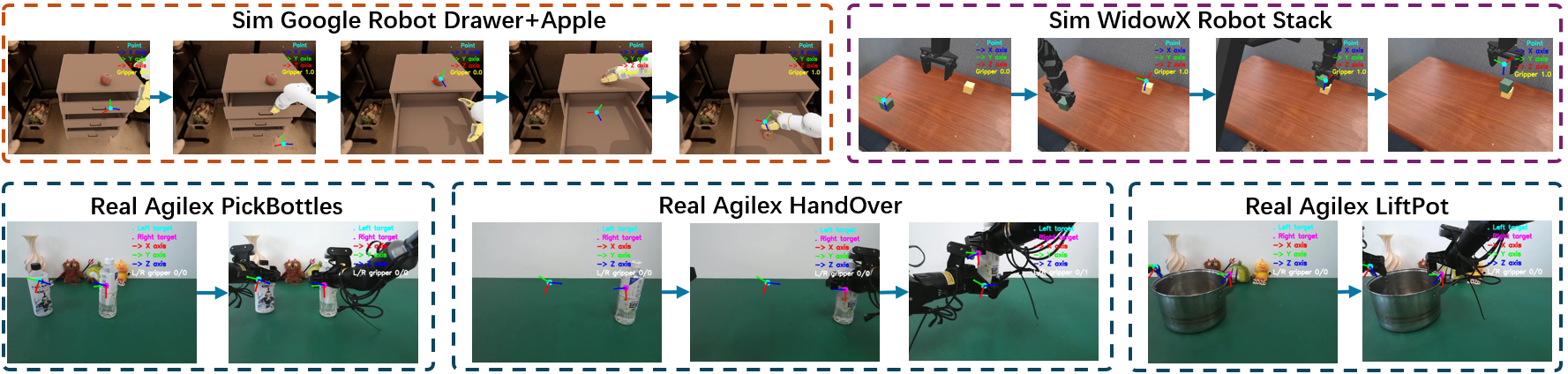}
    \caption{
    Qualitative visualization of predicted key-actions on five representative single-arm and dual-arm simulation and real-robot tasks. H-VLA predicts task-relevant key-actions for diverse manipulation primitives, including picking, opening, placing, and closing.
    }
    \label{fig:keyaction_vis}
\end{figure*}

Fig.~\ref{fig:keyaction_vis} visualizes the predicted key-actions on five representative tasks: \emph{Sim Google Robot Drawer+Apple}, \emph{Sim WidowX Robot Stack}, \emph{Real Agilex PickBottles}, \emph{Real Agilex HandOver}, and \emph{Real Agilex LiftPot}. These examples cover both simulation and real-world settings, as well as single-arm and dual-arm manipulation. The results show that H-VLA can predict task-relevant key-actions for diverse manipulation primitives, including picking, opening, placing, and closing. These structured key-actions provide effective intermediate targets for the Motion Planning Model, enabling it to generate action chunks that complete the corresponding tasks.


\section{Conclusion}
\label{sec:conclusion}

We presented H-VLA, a hierarchical Vision-Language-Action framework that decouples high-level semantic reasoning from low-level motion generation. H-VLA predicts structured key-actions that represent the next manipulation subgoal, generates dense future action chunks with a Motion Planning Model, and represents states, key-actions, and actions in a unified camera-centric action space. Together with a two-stage training strategy, this design improves transfer across heterogeneous datasets, embodiments, and viewpoints. Experiments on SimplerEnv and Agilex show that hierarchical key-action reasoning and camera-centric action representation provide a promising direction for robust and transferable robot manipulation.

\section{Limitations}
\label{sec:limitations}

H-VLA still has several limitations. First, automatic labels based on gripper transitions and trajectory ends can miss salient intermediate contact events. Extending the framework to non-prehensile, deformable-object, or dynamic manipulation requires task-appropriate supervision and further evaluation. Second, the camera-centric representation relies on reliable camera--robot extrinsics. Although our experiments use practical calibration rather than high-precision calibration, the sensitivity to calibration errors has not been systematically evaluated. This requirement does not fundamentally limit extension to humanoid or semi-humanoid robots with head-mounted cameras, where the camera-to-base extrinsics can typically be obtained or updated from the robot's kinematic and calibration information for the coordinate transformation. Third, although states, key-actions, and actions are represented in the third-person camera frame, the VLM mainly provides 2D visual reasoning, while key-actions are 3D targets; connecting 2D semantic grounding with 3D key-action prediction remains important. Fourth, fine-grained dexterous and contact-rich manipulation remains challenging, requiring precise local geometry, force interaction, and contact dynamics.

{
\acknowledgments{
We are grateful to the Samsung AI Center, DS Division, for providing substantial funding and computational resources that supported this research. We also thank our colleagues at Samsung R\&D Institute China-Beijing (SRCB) for helpful discussions and support.
}
}

\bibliography{example}

\clearpage
\appendix
{\LARGE\bfseries Appendix}
\vspace{0.5em}

\section{Dataset Details and Processing}
\label{sec:supp_dataset}

\subsection{Dataset Overview}

H-VLA is trained and evaluated on a mixture of single-arm and dual-arm robot datasets, including RoboCOIN Cobot Magic~\cite{wu2025robocoin}, DROID~\cite{khazatsky2024droid}, BridgeDataV2~\cite{walke2023bridgedata}, FractalData~\cite{brohan2022rt}, and our collected Agilex Real Data. These datasets cover different robot embodiments, manipulation tasks, camera configurations, and calibration settings. Table~\ref{tab:dataset_stats} summarizes the dataset statistics, while Table~\ref{tab:dataset_calibration} summarizes the camera calibration information.

\begin{table*}[h]
\centering
\footnotesize
\setlength{\tabcolsep}{4pt}
\resizebox{\textwidth}{!}{
\begin{tabular}{lccccccc}
\toprule
Dataset & Episodes & Tasks & Embodiment & Arm Type & 3rd-Person & Wrist & Used Episodes \\
\midrule
RoboCOIN Cobot Magic & 20k & 54 & Agilex Robot & Dual-Arm & Yes & Yes & 14k \\
DROID & 76k & 86 & Franka Robot & Single-Arm & Yes & Yes & 22k \\
BridgeDataV2 & 60k & 13 Skills & WidowX Robot & Single-Arm & Yes & No & 60k \\
FractalData & 87k & 700+ & Google Robot & Single-Arm & Yes & No & 87k \\
Agilex Real Data & 150 & 3 & Agilex Robot & Dual-Arm & Yes & Yes & 150 \\
\bottomrule
\end{tabular}
}
\caption{Dataset statistics used in H-VLA. The 3rd-Person and Wrist columns indicate the visual streams used by H-VLA after dataset preprocessing.}
\label{tab:dataset_stats}
\end{table*}

\begin{table*}[h]
\centering
\footnotesize
\setlength{\tabcolsep}{3pt}
\resizebox{\textwidth}{!}{
\begin{tabular}{lccc}
\toprule
Dataset & Camera View / Pose & Intrinsics / Extrinsics & Calibration Method \\
\midrule
RoboCOIN Cobot Magic & Task-dependent & Orbbec factory intrinsics / Self-calibrated extrinsics & Task-specific PnP self-calibration using FK-derived gripper poses and image annotations \\
DROID & Task-dependent & Provided / Provided & Directly obtained from DROID calibration files \\
BridgeDataV2 & Single pose & SimplerEnv / SimplerEnv & SimplerEnv reference calibration; several tasks further corrected using ECoT gripper markers \\
FractalData & Single pose & SimplerEnv / SimplerEnv & Directly obtained from SimplerEnv \\
Agilex Real Data & Fixed pose & Orbbec factory intrinsics / Self-calibrated extrinsics & PnP self-calibration using FK-derived gripper poses and image annotations \\
\bottomrule
\end{tabular}
}
\caption{Camera configurations and calibration sources used by H-VLA after preprocessing.}
\label{tab:dataset_calibration}
\end{table*}

\subsubsection{RoboCOIN Cobot Magic}

RoboCOIN Cobot Magic~\cite{wu2025robocoin} is a dual-arm manipulation dataset collected on an Agilex robot. The full dataset contains about 20k episodes across 54 tasks. It provides third-person and wrist-view observations, but does not directly provide camera intrinsics and extrinsics. We use about 14k processed episodes. For camera intrinsics, we use the Orbbec factory parameters. For camera extrinsics, we calibrate each task using the self-calibration procedure described in Sec.~\ref{sec:supp_self_calib}. Since different tasks may use different camera placements, RoboCOIN contains task-dependent camera poses.

\subsubsection{DROID}

DROID~\cite{khazatsky2024droid} is a large-scale single-arm Franka robot dataset with about 76k episodes across 86 tasks. It provides both third-person and wrist-view observations, together with camera intrinsics and extrinsics. We use about 22k processed episodes in H-VLA. The available calibration enables direct transformation of end-effector states and actions from the robot base frame to the third-person camera frame.

\subsubsection{BridgeDataV2}

BridgeDataV2~\cite{walke2023bridgedata} is a single-arm WidowX robot dataset containing about 60k episodes across 13 skills. H-VLA uses only the primary third-person RGB stream, \texttt{image\_0}; wrist-view and depth inputs are not used. We use the SimplerEnv intrinsics and extrinsics as reference calibration. For several tasks, such as Stack, we further correct the extrinsic parameters using the gripper markers provided by ECoT~\cite{zawalski2024robotic}. We use all 60k episodes in our training pipeline.

\subsubsection{FractalData}

FractalData~\cite{brohan2022rt} is a single-arm Google Robot dataset containing about 87k episodes across more than 700 tasks. It provides third-person observations without wrist-view cameras. Similar to BridgeDataV2, we use the camera intrinsics and extrinsics from SimplerEnv. We use all 87k episodes for training and fine-tuning.

\subsubsection{Agilex Real Data}

Agilex Real Data is our collected dual-arm real-robot dataset. It contains 150 demonstrations across three tasks: \emph{PickBottles}, \emph{HandOver}, and \emph{LiftPot}, with 50 demonstrations per task. The robot is an Agilex dual-arm platform equipped with a fixed third-person camera and two wrist cameras. We use the Orbbec factory intrinsics and estimate the fixed third-person camera extrinsics with the self-calibration method in Sec.~\ref{sec:supp_self_calib}.

\subsection{Dataset Processing Pipeline}
\label{sec:supp_dataset_processing}

To enable joint training across heterogeneous robot embodiments and datasets, all demonstrations are converted into a unified third-person camera-centric action space. This processing pipeline standardizes current end-effector states, key-actions, and action chunks into a common coordinate system aligned with visual observations.

\subsubsection{Coordinate Frames}

We consider two coordinate frames: the robot base frame and the third-person camera frame. Original robot states and actions are typically represented in the robot base frame, while visual observations are captured in the third-person camera frame. H-VLA transforms current end-effector states, key-actions, and actions into the third-person camera frame so that action representations are aligned with visual observations.

\subsubsection{Camera Calibration and Self-Calibration}
\label{sec:supp_self_calib}

For datasets with available camera calibration, such as DROID, we directly use the provided camera intrinsics and extrinsics. For datasets without complete calibration information, including RoboCOIN Cobot Magic and Agilex Real Data, we estimate camera extrinsics using a self-calibration procedure. For RoboCOIN Cobot Magic, the procedure is applied per task because the third-person camera placement may vary across tasks. For Agilex Real Data, the physical third-person camera placement remains fixed across the real-robot experiments, and the same PnP-based procedure is used to obtain the fixed-camera extrinsics. Several gripper positions are manually annotated in the third-person image, and the corresponding 3D end-effector positions are obtained from robot joint states through forward kinematics (FK). The end-effector pose includes the gripper offset used during manipulation. Given the resulting 2D--3D correspondences, camera extrinsics are estimated using a Perspective-n-Point (PnP) solver.
For the primary BridgeDataV2 \texttt{image\_0} stream and for FractalData, camera intrinsics and extrinsics are obtained from SimplerEnv as reference calibration. For several BridgeDataV2 tasks, such as Stack, the extrinsics are further corrected using the gripper markers released by ECoT~\cite{zawalski2024robotic}.

\subsubsection{State Transformation}

For each timestep, the current end-effector pose is transformed from the robot base frame to the third-person camera frame:
\begin{equation}
\mathbf{T}^{c}_{ee}
=
\mathbf{T}^{c}_{b}
\mathbf{T}^{b}_{ee},
\end{equation}
where $\mathbf{T}^{c}_{b}$ denotes the base-to-camera transformation and $\mathbf{T}^{b}_{ee}$ denotes the end-effector pose in the robot base frame. The camera-frame Cartesian position, Euler orientation, and processed gripper state are then extracted as a 7-dimensional vector
\begin{equation}
\mathbf{u}_t=[x,y,z,\phi,\theta,\psi,g]\in\mathbb{R}^{7},
\end{equation}
where the gripper state is binarized such that $g=1$ denotes open and $g=0$ denotes closed. For dual-arm data, the left- and right-arm vectors are concatenated into a 14-dimensional representation, $\mathbf{u}_t=[\mathbf{u}^{L}_t,\mathbf{u}^{R}_t]\in\mathbb{R}^{14}$. Current end-effector states and key-actions are represented as absolute camera-frame poses. Actions use the same dimensionality but encode frame-to-frame delta position and orientation, while the gripper dimension remains an absolute binarized state.

\subsubsection{Key-Action Transformation}

Key-actions are represented as absolute end-effector poses in the third-person camera frame. After key-action labels are extracted from demonstration trajectories, they are transformed using the same base-to-camera transformation and converted to the same 7- or 14-dimensional representation as the current end-effector state. Thus, key-actions from different robot embodiments and datasets are represented in a common visual coordinate frame, with the same binary gripper convention.

\subsubsection{Action Transformation}

Future actions are represented as frame-to-frame delta actions in the third-person camera frame, except for the gripper dimension. We first transform consecutive end-effector poses into the camera frame and convert them to the 7-dimensional representation:
\begin{equation}
\mathbf{u}_{t}, \mathbf{u}_{t+1} \in \mathbb{R}^{7}.
\end{equation}
The single-arm action is then computed as
\begin{equation}
\Delta\mathbf{a}_{t+1}
=
[
\mathbf{p}_{t+1}-\mathbf{p}_{t},
\mathbf{r}_{t+1}-\mathbf{r}_{t},
g_{t+1}
],
\end{equation}
where $\mathbf{p}_t=[x_t,y_t,z_t]$, $\mathbf{r}_t=[\phi_t,\theta_t,\psi_t]$, and the gripper output remains the absolute binarized state at the next timestep rather than a delta. For dual-arm data, the same operation is applied independently to the left and right arms and then concatenated into a 14-dimensional action vector. As a result, action chunks consist of camera-centric delta position and orientation commands with absolute gripper states, while current end-effector states and key-actions remain absolute 7- or 14-dimensional representations in the third-person camera frame.

\subsubsection{Key-Action Label Construction}

Ground-truth key-actions are constructed from gripper-state transitions in demonstration trajectories. For a trajectory of length $T$, let $g_t$ denote the binarized gripper state at timestep $t$, where $g_t=1$ represents an open gripper and $g_t=0$ represents a closed gripper. We define the set of key-action indices as

\begin{equation}
\mathcal{I}^{k}
=
\{\,t+1 \mid |g_{t+1}-g_t| > 0.5,\ 1 \leq t < T\,\}
\cup
\{T\},
\end{equation}

where the frame immediately following a gripper-state transition is treated as a key-action, and the final frame is always included to represent the terminal manipulation state.

For each timestep $t$, the target key-action is defined as the nearest future key-action. Specifically, we first identify the nearest future key-action index

\begin{equation}
\tau(t)
=
\min
\left\{
j \in \mathcal{I}^{k}
\mid
j \ge t
\right\},
\end{equation}

and use the corresponding state as the ground-truth key-action label:

\begin{equation}
\mathbf{k}_t
=
\mathbf{u}_{\tau(t)},
\end{equation}

where $\mathbf{u}_{\tau(t)}$ denotes the absolute end-effector pose and binarized gripper state represented in the unified third-person camera frame.

For dual-arm trajectories, key-actions are extracted independently for the left and right arms using their respective gripper-state transitions. The final dual-arm key-action is represented as

\begin{equation}
\mathbf{k}_t
=
\left[
\mathbf{k}^{L}_t,
\mathbf{k}^{R}_t
\right]
\in
\mathbb{R}^{14},
\end{equation}

where $\mathbf{k}^{L}_t$ and $\mathbf{k}^{R}_t$ denote the nearest future key-actions for the left and right arms, respectively. The left and right key-actions may correspond to different future timesteps, which allows the representation to capture temporally distinct bimanual manipulation stages.

As a result, each timestep is supervised by the nearest future manipulation subgoal, allowing the Key-Action Model to learn semantically meaningful intermediate goals rather than dense low-level actions.

\begin{figure*}[h!]
    \centering
    \includegraphics[width=\textwidth]{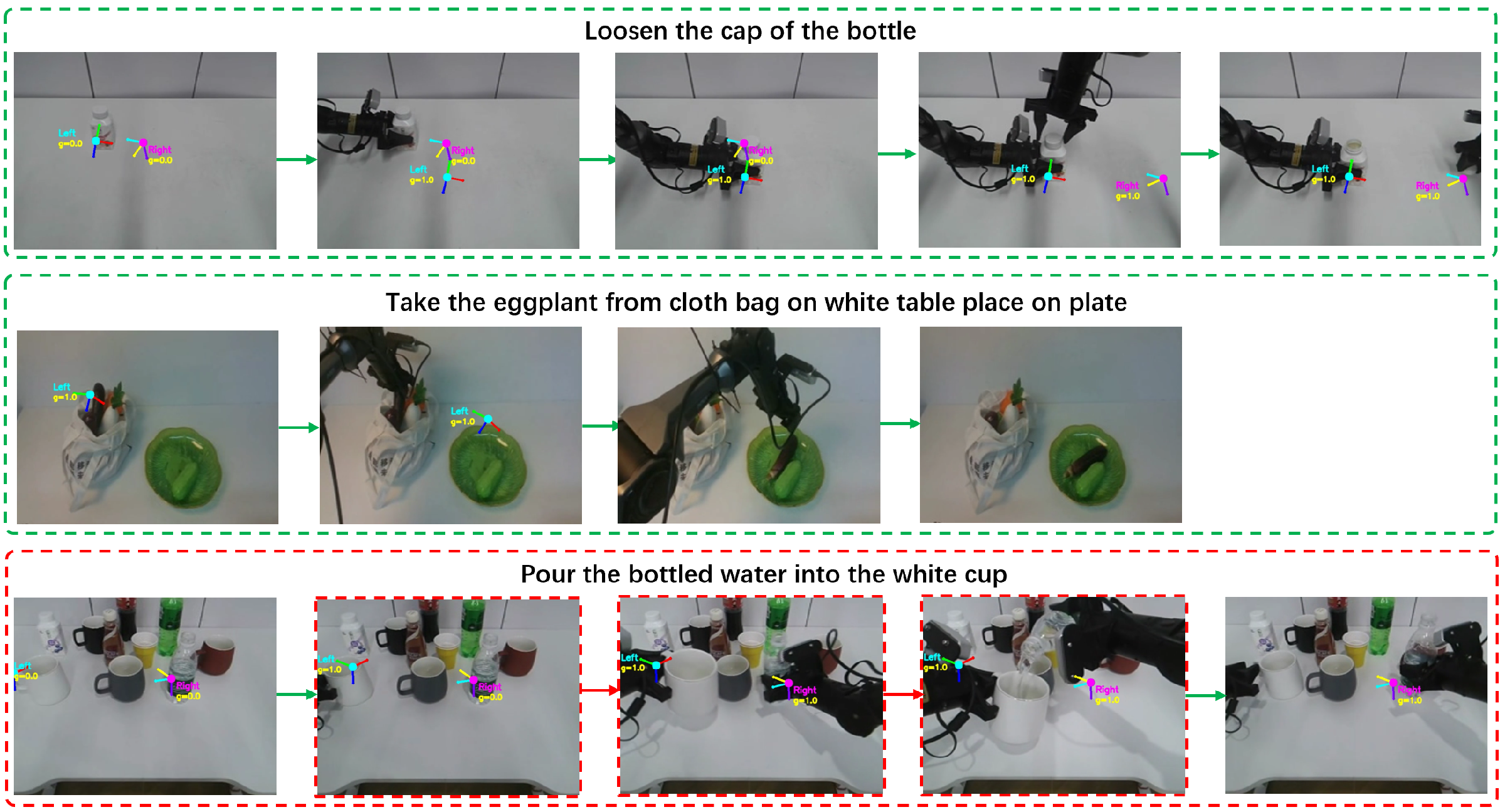}
    \caption{
    Ground-truth key-action visualization on RoboCOIN Cobot Magic. Green dashed boxes indicate correctly identified key-actions, while red dashed boxes and red arrows indicate failure cases.
    }
    \label{fig:robocoin_gt}
\end{figure*}

\begin{figure*}[h!]
    \centering
    \includegraphics[width=\textwidth]{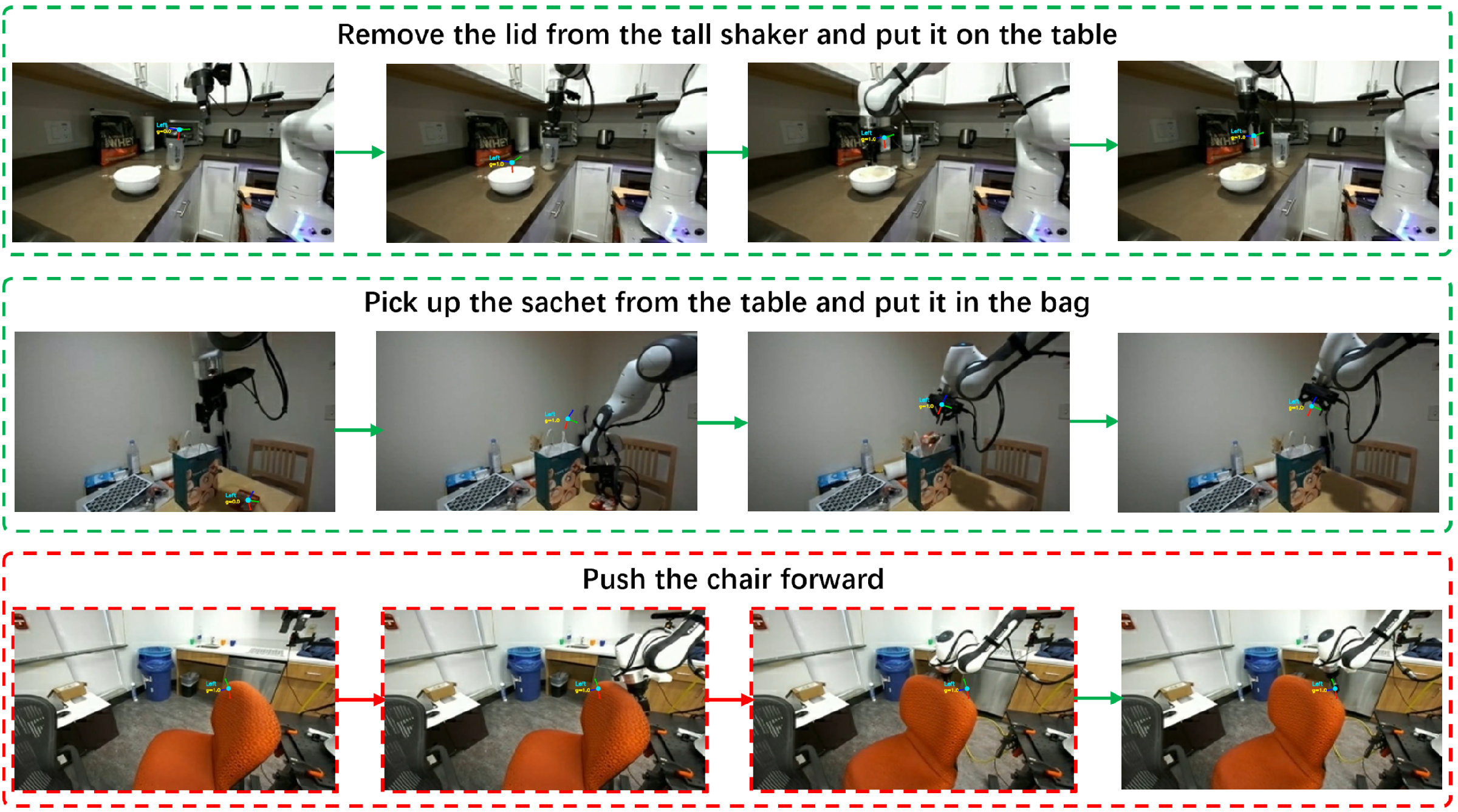}
    \caption{
    Ground-truth key-action visualization on DROID. Green dashed boxes indicate correctly identified key-actions, while red dashed boxes and red arrows indicate failure cases.
    }
    \label{fig:droid_gt}
\end{figure*}

\begin{figure*}[h!]
    \centering
    \includegraphics[width=\textwidth]{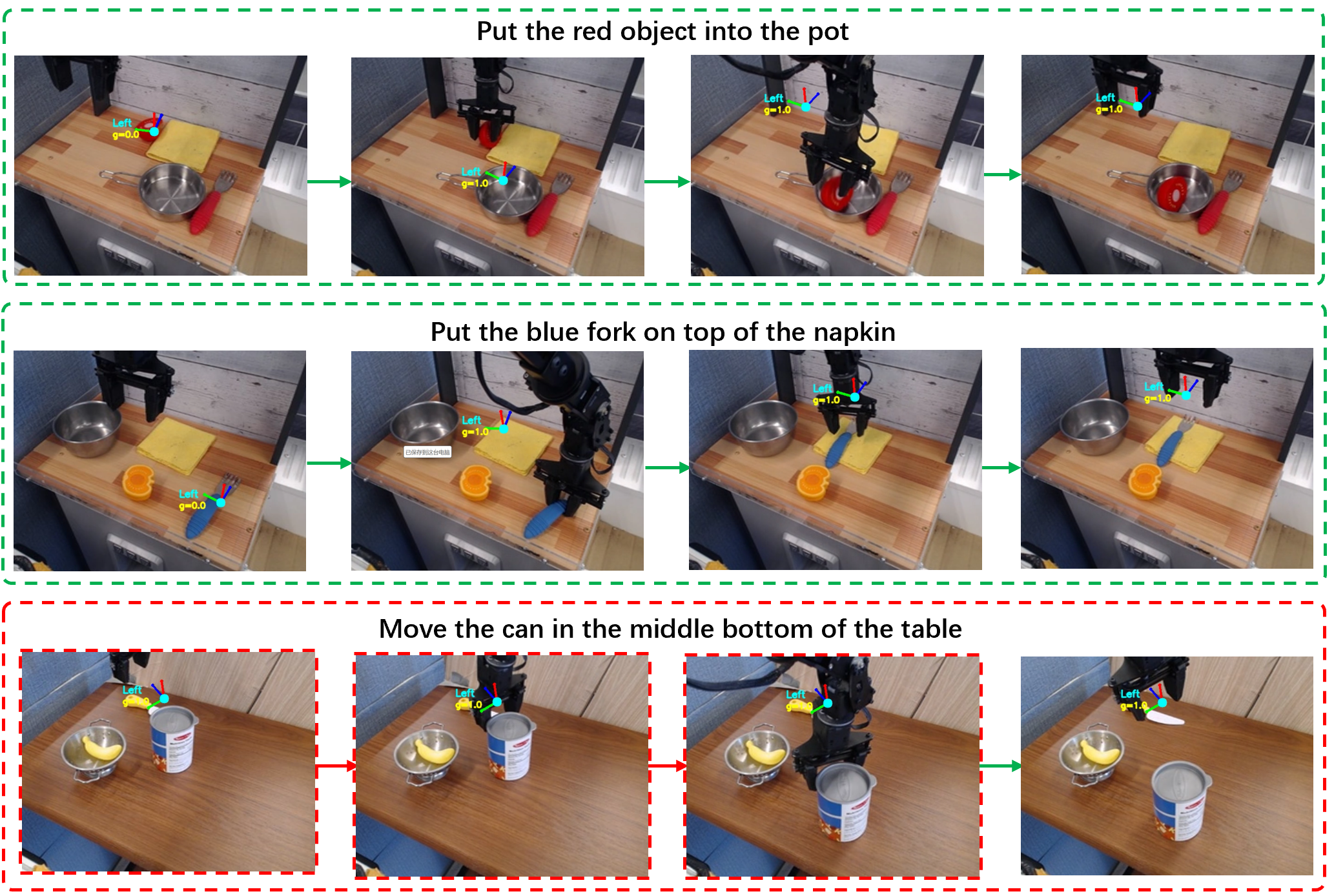}
    \caption{
    Ground-truth key-action visualization on BridgeDataV2. Green dashed boxes indicate correctly identified key-actions, while red dashed boxes and red arrows indicate failure cases.
    }
    \label{fig:bridgedata_gt}
\end{figure*}

\begin{figure*}[h!]
    \centering
    \includegraphics[width=\textwidth]{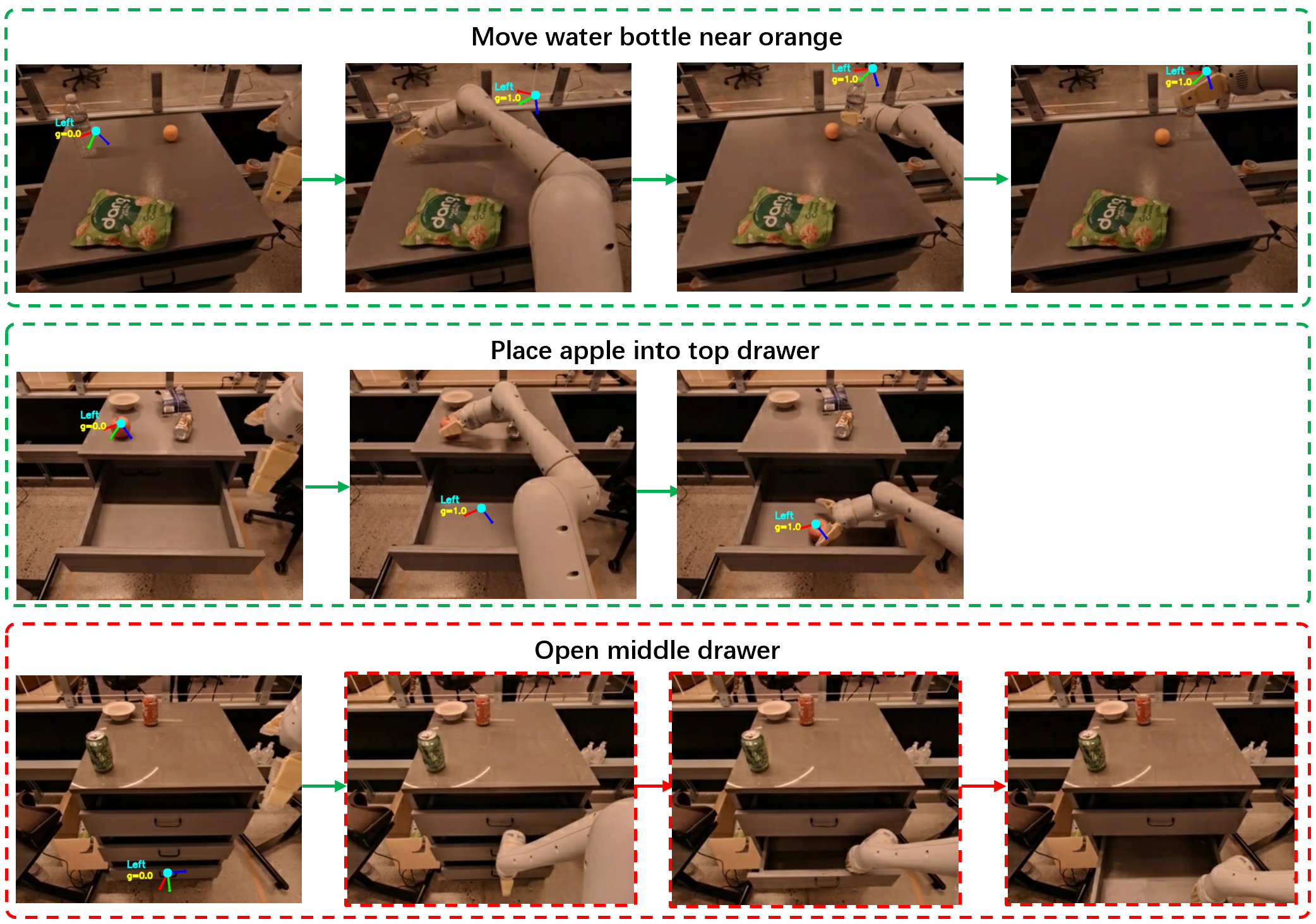}
    \caption{
    Ground-truth key-action visualization on FractalData. Green dashed boxes indicate correctly identified key-actions, while red dashed boxes and red arrows indicate failure cases.
    }
    \label{fig:fractal_gt}
\end{figure*}

\begin{figure*}[h!]
    \centering
    \includegraphics[width=\textwidth]{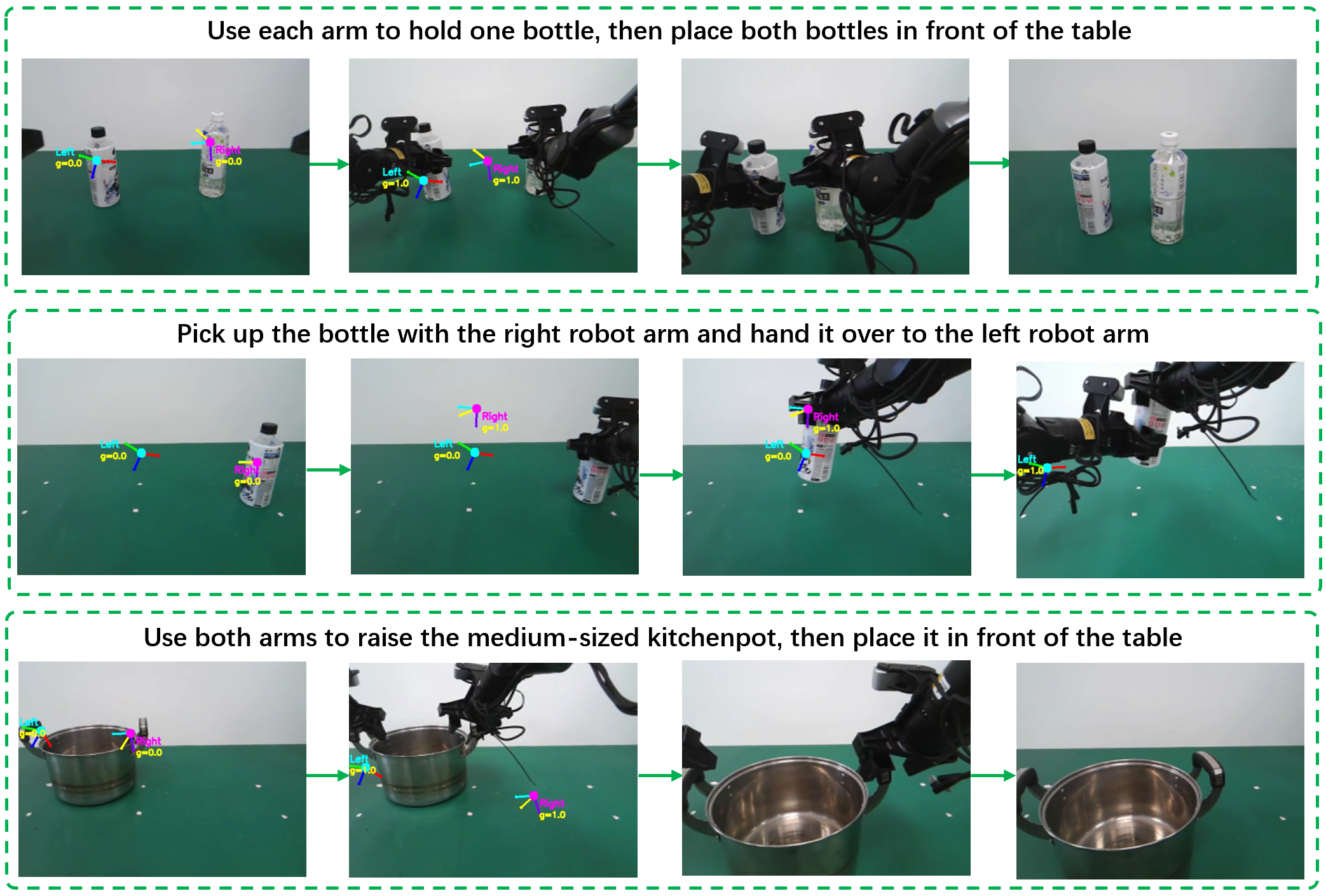}
    \caption{
    Ground-truth key-action visualization on Agilex Real Data. Green dashed boxes indicate correctly identified key-actions, while red dashed boxes and red arrows indicate failure cases.
    }
    \label{fig:agilex_gt}
\end{figure*}

\subsubsection{Ground-Truth Verification and Visualization}

To verify the correctness of camera intrinsic and extrinsic calibration, coordinate transformations, and key-action construction, we project transformed ground-truth key-actions back onto third-person camera images. These visualizations are used to identify calibration errors, inconsistent coordinate transformations, and key-action annotations that are inconsistent with the intended manipulation semantics before training.

We perform this verification procedure on all five datasets used in H-VLA, including RoboCOIN Cobot Magic, DROID, BridgeDataV2, FractalData, and Agilex Real Data. For each dataset, we visualize projected key-actions on representative tasks to ensure that the transformed camera-centric representations are geometrically consistent with the observed manipulation scene.

Figures~\ref{fig:robocoin_gt}--\ref{fig:agilex_gt} present representative visualization examples from the five datasets. In each figure, green dashed boxes indicate examples whose key-action labels are correctly identified by the proposed gripper-based key-action construction pipeline, while red dashed boxes indicate examples whose key-action labels are not correctly identified. For the failure cases, both the outline of the image and the corresponding key-action arrows are highlighted in red.

For the representative tasks involving gripper-state transitions, many projected key-actions visually align with task-relevant grasp, handover, placement, or contact regions. These examples qualitatively illustrate the geometric consistency of the calibration, coordinate transformation, and key-action construction pipeline, while the highlighted failures show its limitations.

\subsubsection{Limitations of Gripper-Based Key-Action Labels}

The current key-action label construction relies on gripper-state transitions and the final trajectory state. This heuristic is effective for many pick-and-place-style manipulation tasks, where opening and closing the gripper naturally indicate important manipulation moments. However, the visualizations in Figures~\ref{fig:robocoin_gt}--\ref{fig:agilex_gt} reveal several failure cases whose critical manipulation states are weakly correlated with gripper-state changes.

For example, in RoboCOIN Cobot Magic, the \emph{Pour Water} task mainly consists of continuous arm motion while the gripper state remains unchanged. In DROID, the \emph{Push Chair Forward} task requires contact-rich pushing rather than grasping. In BridgeDataV2, the \emph{Move Can} task contains important intermediate states that are unrelated to gripper opening or closing. Similarly, in FractalData, the \emph{Open Drawer} task primarily depends on arm motion and contact interaction rather than gripper transitions. Consequently, meaningful manipulation sub-targets may not coincide with gripper-state changes, causing the current heuristic to miss important key-actions.

In the visualizations, these failure cases are highlighted using red dashed boxes and red key-action arrows. These examples reveal a limitation of gripper-based key-action construction. Future work should explore more general key-action annotation or discovery methods, such as trajectory segmentation, contact-state estimation, visual affordance grounding, human-provided keyframe annotations, or learned key-action discovery mechanisms.

\section{Additional Model and Training Details}
\label{sec:supp_model}

\subsection{Overall Architecture}

H-VLA consists of a pre-trained Vision-Language Model (VLM), a Key-Action Model, and a Motion Planning Model. The VLM extracts multimodal cognition features from third-person RGB observations and language instructions. The Key-Action Model predicts a structured key-action representing the next manipulation subgoal, and the Motion Planning Model generates future action chunks conditioned on a key-action, current end-effector state, and multimodal features. Ground-truth key-actions are used as the Motion Planning Model condition during training, whereas Key-Action Model predictions are used at inference.

The 7-dimensional representation denotes a physical single arm, whereas the final H-VLA model uses a shared 14-dimensional interface. For single-arm data, current states, key-actions, and actions are replicated from 7 to 14 dimensions; dual-arm data remain unchanged. The action horizon is controlled by the future action window size. Consistent with the main paper, the action chunk is written as $\mathbf{A}_t=[\Delta\mathbf{a}_{t+1},\Delta\mathbf{a}_{t+2},\dots,\Delta\mathbf{a}_{t+H}]$. In our implementation, \texttt{future\_action\_window\_size}=15, corresponding to $H=16$ future transition commands.

\subsection{Pre-trained Vision-Language Model}

H-VLA uses a Prismatic-7B VLM backbone~\cite{karamcheti2024prismatic}. Third-person RGB observations are encoded by the DINOv2~\cite{oquab2023dinov2} and SigLIP~\cite{zhai2023sigmoid} vision backbones, and language instructions are processed by the LLaMA-2~\cite{touvron2023llama2} language backbone. Following CogACT~\cite{li2024cogact}, a learnable action token is appended to the multimodal token sequence, and its hidden state is used as the cognition feature for downstream key-action prediction and motion planning.

\subsection{Key-Action Model}

{
The Key-Action Model is implemented as a DiT-based diffusion model~\cite{peebles2023scalable} and predicts one structured key-action target. Consistent with the formulation in the main paper, its condition is
\begin{equation}
\mathbf{c}^{k}_t =
\operatorname{concat}
\left(
\mathbf{f}^{cog}_t,
\phi_s(\mathbf{s}_t)
\right),
\end{equation}
where $\mathbf{f}^{cog}_t$ is the VLM cognition feature and $\phi_s(\cdot)$ is a two-layer MLP that encodes the current end-effector state. Let $\mathbf{k}_t$ denote the clean ground-truth key-action and $\mathbf{k}_{t,i}$ its noisy version at diffusion step $i$. The Key-Action Model is trained with
\begin{equation}
\mathcal{L}_{\text{key}} =
\mathbb{E}_{i,\boldsymbol{\epsilon}^{k}}
\left[
\left\|
\epsilon^{k}_{\theta}
(\mathbf{k}_{t,i}, i, \mathbf{c}^{k}_t)
-
\boldsymbol{\epsilon}^{k}
\right\|_2^2
\right].
\end{equation}
For the shared 14-dimensional model interface, the proprioceptive encoder maps the current state from 14 dimensions to a 1024-dimensional feature through a hidden dimension of 2048. In the V1 arm-mask variant described below, a separate arm-mask encoder maps the 14-dimensional mask to a 256-dimensional feature through a hidden dimension of 512. The final H-VLA model uses the V2 single-arm replication strategy and therefore does not contain inactive arm dimensions.
}

\subsection{Motion Planning Model}

{
The Motion Planning Model is also implemented as a DiT-based diffusion model~\cite{peebles2023scalable}. Its condition follows the main-paper formulation:
\begin{equation}
\mathbf{c}^{a}_t =
\operatorname{concat}
\left(
\mathbf{f}^{cog}_t,
\mathbf{f}^{cls}_t,
\phi_w(\mathbf{o}^{wrist}_{t}),
\phi_m(\mathbf{m}^{wrist}_t),
\phi_{ak}\left([\mathbf{s}_t,\mathbf{k}^{\mathrm{cond}}_t]\right)
\right),
\end{equation}
where $\mathbf{f}^{cls}_t$ is the global visual feature, $\phi_w(\cdot)$ encodes wrist-view observations, and $\phi_{ak}(\cdot)$ encodes the current-state--key-action pair. During training, $\mathbf{k}^{\mathrm{cond}}_t$ is the ground-truth key-action; at inference, it is the key-action predicted by the Key-Action Model. For the shared 14-dimensional interface, the concatenated current-state--key-action input has 28 dimensions and is mapped to a 1024-dimensional feature through a hidden dimension of 2048.

\paragraph{Wrist-view conditioning.}
When wrist-view observations are available, left and right wrist images are encoded by the same DINOv2-SigLIP vision backbone~\cite{oquab2023dinov2,zhai2023sigmoid} used by the VLM. The resulting patch features are average-pooled and projected into a 512-dimensional wrist feature. When a wrist view is unavailable, the corresponding feature is zero-filled while preserving the fixed model interface. A wrist-camera mask
\begin{equation}
\mathbf{m}^{wrist} \in \{0,1\}^{2}
\end{equation}
indicates left/right wrist-camera availability. The corresponding wrist features are masked accordingly, and the wrist mask is encoded by a two-layer MLP into a 128-dimensional feature that is appended to the motion-planning condition.

With $H=16$ in our implementation, the clean future action chunk is
\begin{equation}
\mathbf{A}_t =
[\Delta\mathbf{a}_{t+1},
\Delta\mathbf{a}_{t+2},
\dots,
\Delta\mathbf{a}_{t+16}],
\end{equation}
where each action command contains camera-centric delta position and orientation together with the absolute binarized gripper state at the target timestep. Let $\mathbf{A}_{t,i}$ denote the noisy action chunk at diffusion step $i$. The Motion Planning Model is trained with
\begin{equation}
\mathcal{L}_{\text{act}} =
\mathbb{E}_{i,\boldsymbol{\epsilon}^{a}}
\left[
\left\|
\epsilon^{a}_{\theta}
(\mathbf{A}_{t,i}, i, \mathbf{c}^{a}_t)
-
\boldsymbol{\epsilon}^{a}
\right\|_2^2
\right].
\end{equation}
In the V1 arm-mask variant, the validity mask for the current-state--key-action pair is also encoded into a 256-dimensional conditioning feature.
}

\subsection{Mixed Single-/Dual-Arm Representation}

To support mixed single-arm and dual-arm training, H-VLA uses a shared 14-dimensional model interface for current states, key-actions, and actions. We explored three strategies for mapping single-arm data into this interface. The final H-VLA models reported in the main paper use V2 single-arm replication; arm-mask conditioning and masked diffusion operations are used only in the V1 variant.

\paragraph{V1: Arm-Mask Conditioning}

In V1, single-arm demonstrations occupy one valid 7-dimensional arm block within the 14-dimensional interface, while the other arm block is inactive; dual-arm demonstrations use both arm blocks. A 14-dimensional arm-validity mask
\begin{equation}
\mathbf{m}^{arm} \in \{0,1\}^{14}
\end{equation}
identifies active dimensions for the current state, key-action, and action representations. During training, inactive dimensions are excluded from both key-action and action diffusion supervision, and the mask is encoded as additional conditioning information for the corresponding Key-Action and Motion Planning modules. During inference, the same validity mask suppresses predictions associated with inactive arm dimensions.

\paragraph{V2: Single-Arm Replication (Final Version)}

For single-arm demonstrations, the 7-dimensional current state, key-action, and action are independently copied into both arm branches of the shared 14-dimensional interface:
{
\begin{equation}
\mathbf{s}_t \rightarrow [\mathbf{s}_t,\mathbf{s}_t],\qquad
\mathbf{k}_t \rightarrow [\mathbf{k}_t,\mathbf{k}_t],\qquad
\Delta\mathbf{a}_t \rightarrow [\Delta\mathbf{a}_t,\Delta\mathbf{a}_t].
\end{equation}
}
Dual-arm demonstrations remain unchanged. Consequently, all demonstrations use fully populated 14-dimensional state, key-action, and action representations without inactive arm dimensions.

\paragraph{V3: Zero Padding}

For single-arm demonstrations, the current state, key-action, and action are extended to 14 dimensions by zero-padding the inactive arm block:
{
\begin{equation}
\mathbf{s}_t \rightarrow [\mathbf{s}_t,\mathbf{0}],\qquad
\mathbf{k}_t \rightarrow [\mathbf{k}_t,\mathbf{0}],\qquad
\Delta\mathbf{a}_t \rightarrow [\Delta\mathbf{a}_t,\mathbf{0}],
\end{equation}
where $\mathbf{0}\in\mathbb{R}^{7}$.
}
Dual-arm demonstrations remain unchanged.

\paragraph{Experimental Comparison}

We experimentally evaluated all three strategies during mixed single-arm and dual-arm pre-training. Although V1 and V3 explicitly preserve the distinction between single-arm and dual-arm embodiments, both lead to a noticeable performance drop on Google Robot tasks in SimplerEnv after mixed pre-training. We hypothesize that the inactive dimensions introduced by masking or zero-padding create a distribution mismatch between single-arm and dual-arm demonstrations and reduce the effectiveness of shared representation learning.

In our exploratory comparison, V2 performs best overall across the evaluated downstream tasks among the three strategies. By using a fully populated shared 14-dimensional interface for single-arm states, key-actions, and actions, all datasets provide supervision to the same model dimensions for both key-action prediction and motion planning. Therefore, all H-VLA models reported in the main paper use the V2 strategy.

\subsection{Training Objective}

The training objective combines key-action prediction and motion planning:
\begin{equation}
\mathcal{L}
=
\lambda_k \mathcal{L}_{\text{key}}
+
\lambda_a \mathcal{L}_{\text{act}}.
\end{equation}
In Stage 1 Key-Action-Centric Pre-training, we set
\begin{equation}
\lambda_k=10,
\qquad
\lambda_a=1.
\end{equation}
This corresponds to the implementation parameter \texttt{key\_action\_loss\_para}=10. In Stage 2 Motion-Planning-Oriented Fine-tuning, we set $\lambda_k=\lambda_a=1$, retaining both key-action and action supervision while adapting the model to downstream task data.

All model parameters are updated during both pre-training and fine-tuning in our experiments, including the VLM, Key-Action Model, Motion Planning Model, proprioceptive encoders, wrist-image projector, and wrist-mask encoder when wrist-view inputs are enabled. Arm-mask encoders are updated only in the V1 arm-mask conditioning variant.

\subsection{Pre-training Details}

H-VLA is pre-trained on a mixed set of single-arm and dual-arm datasets, including RoboCOIN Cobot Magic~\cite{wu2025robocoin}, DROID~\cite{khazatsky2024droid}, BridgeDataV2~\cite{walke2023bridgedata}, and FractalData~\cite{brohan2022rt}. The pre-training stage uses 8 NVIDIA H100 GPUs for approximately three days and one epoch. Unless otherwise specified, pre-training uses a constant learning rate of $2\times10^{-5}$ and a batch size of 256.

The goal of pre-training is to learn transferable key-action reasoning and camera-centric motion priors from heterogeneous robot embodiments, camera viewpoints, and task distributions.

\subsection{Fine-tuning Details}

For simulation transfer, H-VLA is separately fine-tuned on FractalData~\cite{brohan2022rt} and BridgeDataV2~\cite{walke2023bridgedata}. FractalData fine-tuning is performed on 8 NVIDIA H100 GPUs for approximately three days and four epochs. BridgeDataV2 fine-tuning is performed on 8 NVIDIA H100 GPUs for approximately three days and six epochs. The resulting policies are evaluated on SimplerEnv.

For real-robot adaptation, H-VLA is fine-tuned on 150 Agilex demonstrations across three dual-arm tasks: \emph{PickBottles}, \emph{HandOver}, and \emph{LiftPot}. The Agilex model is fine-tuned using 4 NVIDIA H100 GPUs for approximately one day and 100 epochs.

\section{SimplerEnv Simulation Setup}
\label{sec:supp_simplerenv}

\subsection{Benchmark Overview}

We evaluate H-VLA on SimplerEnv~\cite{li2024evaluating}, including the Google Robot and WidowX Robot simulation benchmarks. For Google Robot, we report both Visual Matching (VM) and Variant Aggregation (VA). For WidowX Robot, we report Visual Matching (VM).

\subsection{SimplerEnv Evaluation Setup}

For readability, the main paper uses shortened task names. Table~\ref{tab:simplerenv_default_protocol} summarizes the task-name mapping and evaluation trial counts. These counts follow the task loops in the released SimplerEnv evaluation scripts~\cite{li2024evaluating}, rather than a new trial-count protocol introduced by H-VLA. Google Robot uses 864 VM and 1992 VA trials (2856 per model/checkpoint), while WidowX uses 24 trials per task (96 across four tasks). Under this setup, environment success does not itself trigger early episode termination; evaluation uses the task-specific maximum horizons configured by the released scripts.

\begin{table}[h]
\centering
\small
\setlength{\tabcolsep}{3pt}
\begin{tabular}{llcc}
\toprule
Short Name & Full Task Name & VM & VA \\
\midrule
Coke Can & Pick Coke Can & 300 & 825 \\
Move Near & Move Near & 240 & 600 \\
Open/Close & Open/Close Drawer & 216 & 378 \\
Drawer+Apple & Open Top Drawer and Place Apple & 108 & 189 \\
Spoon Towel & Put Spoon on Towel & 24 & -- \\
Carrot Plate & Put Carrot on Plate & 24 & -- \\
Stack & Stack Green Block on Yellow Block & 24 & -- \\
Eggplant Basket & Put Eggplant in Yellow Basket & 24 & -- \\
\bottomrule
\end{tabular}
\caption{
Task-name mapping and number of evaluation trials in the SimplerEnv benchmark sweep used for our evaluations. Google Robot tasks are evaluated under both VM and VA settings, while WidowX Robot tasks are evaluated under VM only.
}
\label{tab:simplerenv_default_protocol}
\end{table}

Google Robot tasks use task-specific episode lengths: 80 steps for \emph{Pick Coke Can}, 80 steps for \emph{Move Near}, 113 steps for \emph{Open/Close Drawer}, and 200 steps for \emph{Open Top Drawer and Place Apple}. WidowX Robot tasks use 60 steps.

\subsection{Comparison with X-VLA Modified Setup}

Compared with the released SimplerEnv setup, the X-VLA~\cite{zheng2025x} configuration mainly differs in \textbf{(1) maximum episode horizon}, \textbf{(2) success-triggered early termination}, and \textbf{(3) camera extrinsics} for the WidowX Eggplant Basket task. It also changes the Open/Close Drawer VA trial count from 378 to 210. For a matched comparison, we evaluate X-VLA under the same SimplerEnv setup used for our reported results.

\subsubsection{Google Robot}

Table~\ref{tab:xvla_google_detail} compares the SimplerEnv evaluation setup with the X-VLA-modified configuration on Google Robot tasks.

\begin{table}[h]
\centering
\small
\setlength{\tabcolsep}{3pt}
\resizebox{\linewidth}{!}{
\begin{tabular}{lcccccc}
\toprule
Task & SimplerEnv Success Stop & X-VLA Success Stop & SimplerEnv Steps & X-VLA Steps & SimplerEnv Trials & X-VLA Trials \\
\midrule
Coke Can & No & Yes & 80 & $80\times2$ & VM(300)+VA(825) & VM(300)+VA(825) \\
Move Near & No & Yes & 80 & $80\times2$ & VM(240)+VA(600) & VM(240)+VA(600) \\
Open/Close & No & Yes & 113 & $113\times2$ & VM(216)+VA(378) & VM(216)+VA(210) \\
Drawer+Apple & No & Yes & 200 & $200\times2$ & VM(108)+VA(189) & VM(108)+VA(189) \\
\bottomrule
\end{tabular}}
\caption{
Comparison between the SimplerEnv evaluation setup and the X-VLA-modified configuration on Google Robot tasks. X-VLA enables success-triggered early termination, doubles the maximum episode horizon, and changes the Open/Close Drawer VA trial count from 378 to 210.
}
\label{tab:xvla_google_detail}
\end{table}

For Google Robot, X-VLA enables success-triggered early termination, doubles the task-specific maximum horizons, and reduces the Open/Close Drawer VA sweep from 378 to 210 trials.

\subsubsection{WidowX Robot}

Table~\ref{tab:xvla_widowx_detail} compares the SimplerEnv evaluation setup with the X-VLA-modified configuration on WidowX Robot tasks.

\begin{table}[h]
\centering
\small
\setlength{\tabcolsep}{3pt}
\resizebox{\linewidth}{!}{
\begin{tabular}{lccccc}
\toprule
Task & SimplerEnv Success Stop & X-VLA Success Stop & SimplerEnv Steps & X-VLA Steps & Camera Transform \\
\midrule
Spoon Towel & No & Yes & 60 & 1200 & Default \\
Carrot Plate & No & Yes & 60 & 1200 & Default \\
Stack & No & Yes & 60 & 1200 & Default \\
Eggplant Basket & No & Yes & 60 & 1200 & Modified by X-VLA \\
\bottomrule
\end{tabular}}
\caption{
Comparison between the SimplerEnv evaluation setup and the X-VLA-modified configuration on WidowX Robot tasks. X-VLA enables success-triggered early termination, increases the maximum episode horizon from 60 to 1200 steps, and modifies the robot-base-to-camera transform for Eggplant Basket.
}
\label{tab:xvla_widowx_detail}
\end{table}

For WidowX Robot, X-VLA enables success-triggered early termination, increases the maximum horizon from 60 to 1200 steps, and modifies the robot-base-to-camera extrinsics for \emph{Put Eggplant in Yellow Basket}.

All H-VLA results and the matched X-VLA comparison in Table~\ref{tab:simplerenv_all} use the SimplerEnv evaluation setup described above.

\subsection{Key-Action Visualization}
\label{sec:supp_visualization}

\subsubsection{Google Robot}

\begin{figure}[h!]
\centering
\includegraphics[width=1\linewidth]{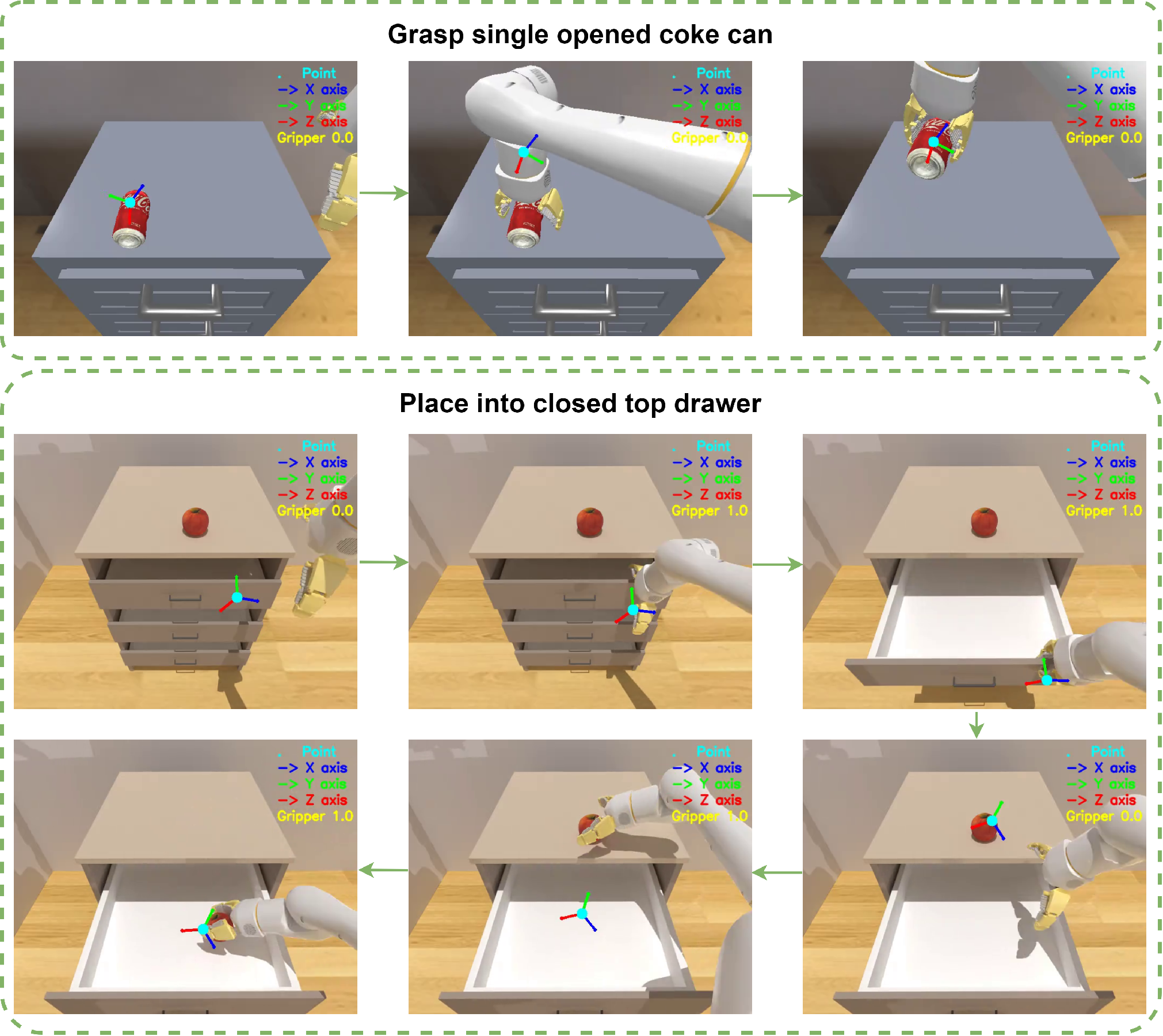}
\caption{
Key-action visualization on Google Robot. We select one short-horizon task and one long-horizon task to illustrate the predicted key-actions at different task phases.
}
\label{fig:google_inference_key_vis}
\end{figure}

In Fig.~\ref{fig:google_inference_key_vis}, we visualize the predicted key-actions on both short-horizon and long-horizon Google Robot tasks. Across the representative task phases shown, the predicted key-actions are visually aligned with task-relevant objects, grasping regions, and placement targets. The long-horizon examples further illustrate that the Key-Action Model can provide task-relevant intermediate targets at multiple manipulation stages rather than only near episode completion, supplying structured conditioning for downstream motion generation.

\subsubsection{WidowX Robot}

\begin{figure}[h!]
\centering
\includegraphics[width=1\linewidth]{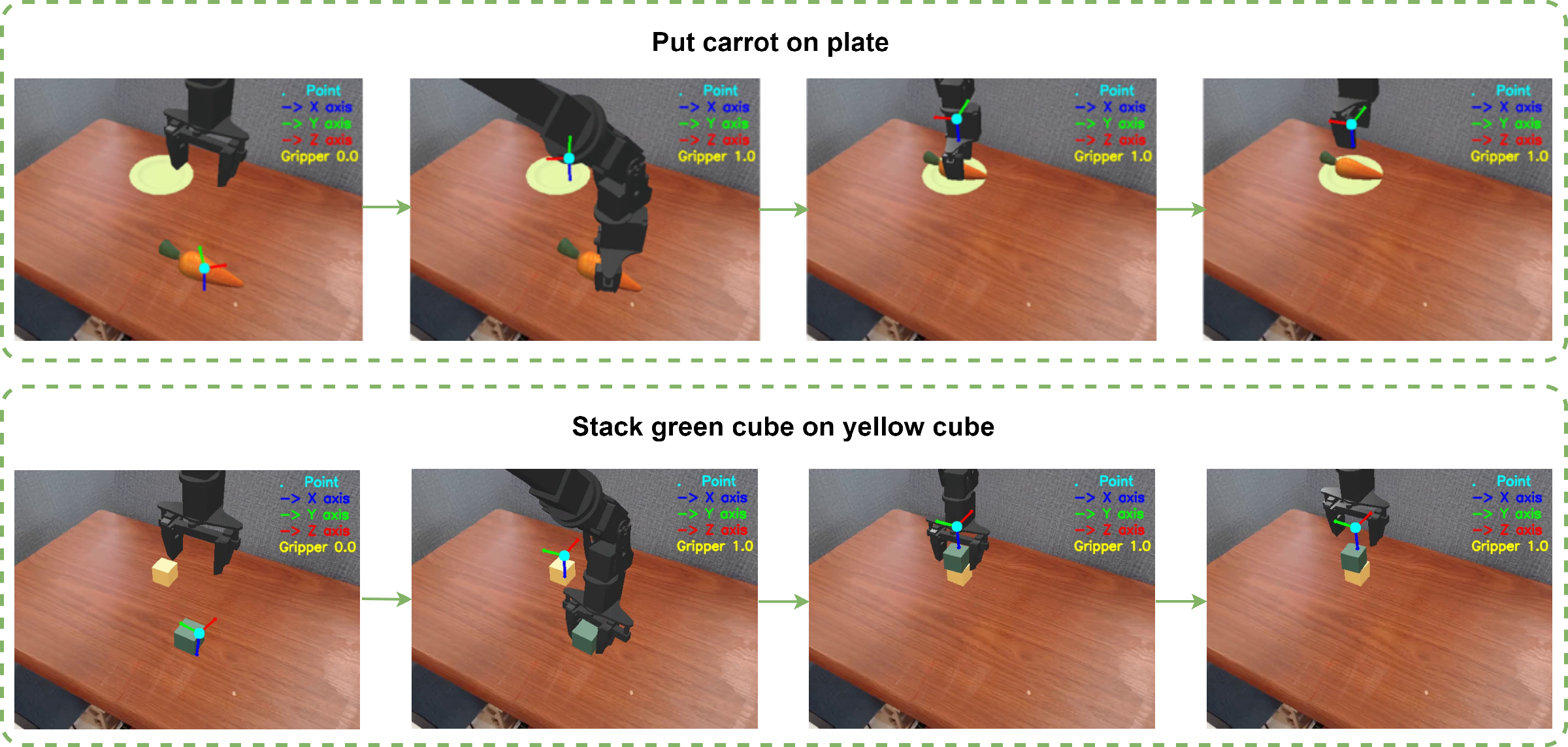}
\caption{
Key-action visualization on WidowX Robot. The cube stacking task requires high precision and fine-grained spatial reasoning.
}
\label{fig:widowx_inference_key_vis}
\end{figure}

In Fig.~\ref{fig:widowx_inference_key_vis}, we visualize the predicted key-actions on two WidowX Robot tasks. Among the evaluated tasks, cube stacking presents a distinct challenge because imprecise placement can easily cause the cubes to fall and lead to task failure. In the representative examples shown, the predicted key-actions are located near task-relevant grasp and placement targets at different manipulation stages, including the precise placement phase required for stacking. These visualizations qualitatively illustrate how the Key-Action Model provides spatially structured intermediate targets for fine-grained manipulation.

\section{Agilex Real-Robot Setup}
\label{sec:supp_real}

\subsection{Hardware Platform}

The real-robot experiments are conducted on an Agilex dual-arm manipulation platform based on a Cobot Magic-style configuration.
The platform consists of two follower robot arms for policy execution and two leader arms for teleoperated demonstration collection.
During data collection, a human operator manipulates the leader arms, and the follower arms reproduce the corresponding motions.
During policy evaluation, only the two follower arms are controlled by the learned policy.

Each manipulation arm is an Agilex PiPER robotic arm.
PiPER is a lightweight 6-DoF robotic arm equipped with an integrated controller and a parallel gripper.
This configuration is suitable for tabletop dual-arm manipulation tasks that require grasping, object transfer, and coordinated bimanual lifting.

The perception system consists of three Orbbec-DABAI RGB-D cameras: one third-person camera and two wrist-mounted cameras.
The third-person camera is mounted in front of the tabletop workspace to observe the global scene, while the two wrist cameras are attached to the left and right follower arms to provide local visual observations for fine-grained manipulation.
Although the cameras provide RGB-D streams, all evaluated policies use RGB observations as visual inputs.

\begin{figure*}[h!]
    \centering
    \includegraphics[width=\textwidth]{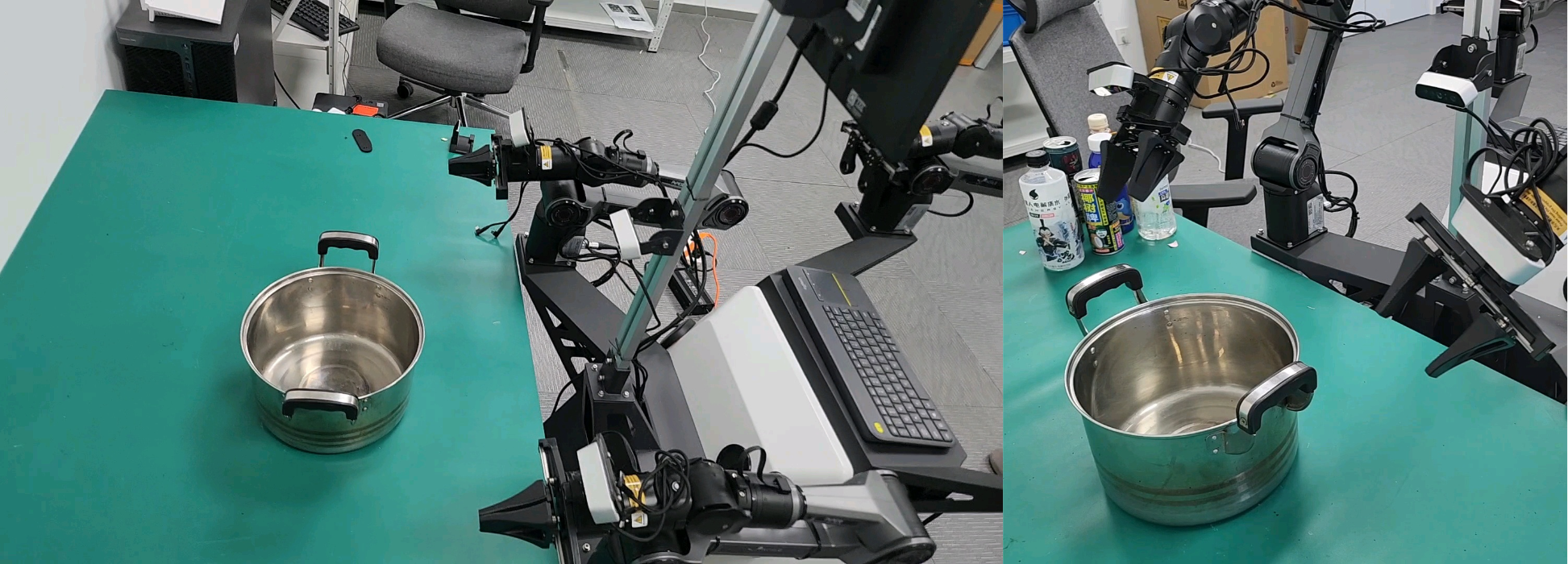}
    \caption{
    Agilex dual-arm robot platform used in the real-world experiments.
    The setup consists of two follower robot arms, two leader arms for teleoperation, one third-person RGB camera, two wrist-mounted RGB cameras, and a tabletop manipulation workspace.
    }
    \label{fig:agilex_setup}
\end{figure*}

All methods are evaluated using the same robot hardware, camera placement, task setup, and success criteria. H-VLA, $\pi_0$, $\pi_{0.5}$, and MolmoAct2 use one third-person and two wrist views, while CogACT uses the third-person view only.
Policy inference is executed on a workstation equipped with a single NVIDIA RTX 4090 GPU.
The hardware configuration is summarized in Table~\ref{tab:agilex_hardware}, and the physical real-robot setup is shown in Fig.~\ref{fig:agilex_setup}.

\begin{table}[h!]
\centering
\small
\setlength{\tabcolsep}{4pt}
\begin{tabular}{lll}
\toprule
Component & Model / Configuration & Description \\
\midrule
Robot platform & Agilex Cobot Magic-style platform & Dual-arm tabletop manipulation setup \\
Manipulation arms & Agilex PiPER arms & Two 6-DoF follower arms for policy execution \\
Teleoperation arms & Agilex PiPER arms & Two leader arms for demonstration collection \\
End-effectors & Parallel grippers & Used for grasping, handover, and lifting \\
Third-person camera & Orbbec-DABAI RGB-D camera & Global workspace observation \\
Wrist cameras & Two Orbbec-DABAI RGB-D cameras & Local left/right arm observations \\
Inference GPU & NVIDIA RTX 4090 & Used for all policy inference experiments \\
\bottomrule
\end{tabular}
\caption{
Hardware configuration of the Agilex real-robot evaluation platform.
}
\label{tab:agilex_hardware}
\end{table}

\subsection{Scene Layout and Tasks}

All tasks are performed on a tabletop workspace observed by the third-person camera, as illustrated in Fig.~\ref{fig:agilex_setup}.
Objects are placed within the reachable workspace of both follower arms.
The third-person camera provides a global view of the scene layout and object positions, while the wrist cameras provide local observations for close-range grasping and manipulation.

We evaluate all methods on three representative dual-arm manipulation tasks: PickBottles, HandOver, and LiftPot.
These tasks cover simultaneous dual-arm manipulation, sequential inter-arm object transfer, and coordinated bimanual lifting.
For training, H-VLA, $\pi_0$~\cite{black2024pi_0}, $\pi_{0.5}$~\cite{intelligence2025pi_5}, CogACT~\cite{li2024cogact}, and MolmoAct2~\cite{fang2026molmoact2} are all fine-tuned on the same 150-demonstration Agilex dataset, producing one unified policy per method for all three tasks. H-VLA is initialized from the mixed pre-trained model, $\pi_0$ and $\pi_{0.5}$ from the OpenPI base checkpoints \texttt{pi0\_base} and \texttt{pi05\_base}, respectively, CogACT from its OXE-pretrained checkpoint, and MolmoAct2 from the MolmoAct2-BimanualYAM checkpoint.

\subsubsection{PickBottles}

\textbf{Language instruction:}

\emph{Use each arm to hold one bottle, then place both bottles in the front area of the table.}

This task requires the robot to use both arms simultaneously.
Each arm must approach and hold one bottle, and the two arms then move the bottles to the front area of the table according to the language instruction.
Successful completion requires accurate grounding of the two bottle targets, correct arm-object association, and spatially decoupled dual-arm motion so that each arm manipulates its corresponding bottle without interfering with the other arm.

\subsubsection{HandOver}

\textbf{Language instruction:}

\emph{Pick up the bottle with the right robot arm and hand it over to the left robot arm.}

This task requires a sequential inter-arm object transfer from the right arm to the left arm.
The right arm first approaches and picks up the bottle, then moves it to a suitable handover pose.
The left arm must coordinate with the right arm to receive the bottle and complete the handover.
This task emphasizes temporal coordination between the two arms, stable object transfer, and accurate reasoning over the right-to-left handover process specified by the language instruction.

\subsubsection{LiftPot}

\textbf{Language instruction:}

\emph{Use both arms to raise the medium-sized kitchen pot, then place it in the front area of the table.}

This task requires both robot arms to jointly grasp and raise a medium-sized kitchen pot, then place it in the front area of the table.
Unlike PickBottles, where the two arms manipulate separate objects, LiftPot requires both arms to manipulate a shared object.
Successful execution therefore requires synchronized bimanual motion, stable two-arm grasping, and coordinated placement to keep the pot balanced throughout the lifting and placement process.

\subsection{Evaluation Settings}

All methods use the same task definitions and success criteria; trial counts follow the evaluation configurations summarized below.
We consider three evaluation settings: In-Distribution (ID), OOD Position, and OOD Scene/Object.
ID evaluates the policy under object placements and scene layouts similar to those observed during training.
OOD Position mainly emphasizes the diversity of position variations: the manipulated objects are placed at diverse unseen positions within the workspace while the task setup and object categories remain unchanged.
OOD Scene/Object introduces broader visual and semantic variations, including changes in background layouts, distractor objects, novel manipulated objects, and mixed variations that combine these factors. All five methods are evaluated under ID and OOD Position, while OOD Scene/Object is evaluated for H-VLA, $\pi_0$, and $\pi_{0.5}$.

\subsubsection{Evaluation Configurations}

\paragraph{In-Distribution (ID).}

Objects are placed at positions similar to those observed during training.
The scene layout, camera configuration, and manipulated objects remain unchanged.
This setting evaluates whether the policy can execute the learned tasks under familiar visual and spatial conditions.

\paragraph{OOD Position.}

The manipulated objects are moved to diverse unseen positions within the reachable workspace while keeping the scene layout and object categories unchanged.
This setting evaluates spatial generalization under diverse position changes, including whether the policy can adapt to varied object placements and still localize, grasp, transfer, lift, or place the target objects correctly.

\paragraph{OOD Scene/Object.}

The scene appearance, distractor configuration, and manipulated objects are modified.
This setting includes changes in background layouts, the introduction of distractor objects, novel target objects to be grasped or manipulated, and mixed variations that combine background changes, distractors, and new object instances.
It evaluates generalization to unseen visual environments, object appearances, and more cluttered or visually different manipulation scenes.

\subsubsection{Evaluation Trial Counts}

H-VLA, $\pi_0$, and $\pi_{0.5}$ are evaluated with 20 trials per task under ID, 30 trials per task under OOD Position, and 30/30/18 trials for PickBottles/HandOver/LiftPot under OOD Scene/Object, yielding 228 trials per model. CogACT and MolmoAct2 are evaluated under ID and OOD Position with 15/15/10 trials for PickBottles/HandOver/LiftPot, yielding 80 trials per model; OOD Scene/Object is not evaluated for these two baselines.

\begin{table}[h!]
\centering
\small
\setlength{\tabcolsep}{4pt}
\begin{tabular}{llcccc}
\toprule
Models & Setting & PickBottles & HandOver & LiftPot & Total \\
\midrule
\multirow{3}{*}{H-VLA / $\pi_0$ / $\pi_{0.5}$}
& ID & 20 & 20 & 20 & 60 \\
& OOD Position & 30 & 30 & 30 & 90 \\
& OOD Scene/Object & 30 & 30 & 18 & 78 \\
\midrule
\multirow{3}{*}{CogACT / MolmoAct2}
& ID & 15 & 15 & 10 & 40 \\
& OOD Position & 15 & 15 & 10 & 40 \\
& OOD Scene/Object & -- & -- & -- & -- \\
\bottomrule
\end{tabular}
\caption{
Number of real-robot evaluation trials per model and setting.
}
\label{tab:agilex_trials}
\end{table}

\begin{table*}[h!]
\centering
\small
\setlength{\tabcolsep}{3pt}
\resizebox{\textwidth}{!}{
\begin{tabular}{lcccc|cccc|cccc}
\toprule
& \multicolumn{4}{c|}{ID} & \multicolumn{4}{c|}{OOD Position} & \multicolumn{4}{c}{OOD Scene/Object} \\
\cmidrule(lr){2-5}\cmidrule(lr){6-9}\cmidrule(lr){10-13}
Policy & PickBottles & HandOver & LiftPot & Avg & PickBottles & HandOver & LiftPot & Avg & PickBottles & HandOver & LiftPot & Avg \\
\midrule
CogACT~\cite{li2024cogact}
& 7/15 & 8/15 & 5/10 & 50\%
& 3/15 & 5/15 & 2/10 & 24\%
& -- & -- & -- & -- \\
MolmoAct2~\cite{fang2026molmoact2}
& 12/15 & 9/15 & 7/10 & 70\%
& 8/15 & 2/15 & 4/10 & 36\%
& -- & -- & -- & -- \\
$\pi_0$~\cite{black2024pi_0}
& 12/20 & \textbf{20/20} & \textbf{18/20} & 83\%
& 10/30 & 9/30 & 9/30 & 31\%
& 11/30 & 7/30 & \textbf{13/18} & 44\% \\
$\pi_{0.5}$~\cite{intelligence2025pi_5}
& 17/20 & 16/20 & 16/20 & 82\%
& 7/30 & 11/30 & 10/30 & 31\%
& 14/30 & 11/30 & 12/18 & 50\% \\
\textbf{H-VLA}
& \textbf{20/20} & 18/20 & \textbf{18/20} & \textbf{93\%}
& \textbf{23/30} & \textbf{28/30} & \textbf{24/30} & \textbf{83\%}
& \textbf{22/30} & \textbf{19/30} & 11/18 & \textbf{66\%} \\
\bottomrule
\end{tabular}}
\caption{
Success counts on the Agilex real-robot benchmark. Each entry reports successful trials over total trials. The average is the mean of the three task success rates, matching the main paper. Dashes indicate settings that were not evaluated. The best result in each column is shown in bold.
}
\label{tab:agilex_success_counts}
\end{table*}

\subsection{Key-Action Visualization}
\label{sec:supp_real_key_visualization}

\begin{figure}
    \centering
    \includegraphics[width=1\linewidth]{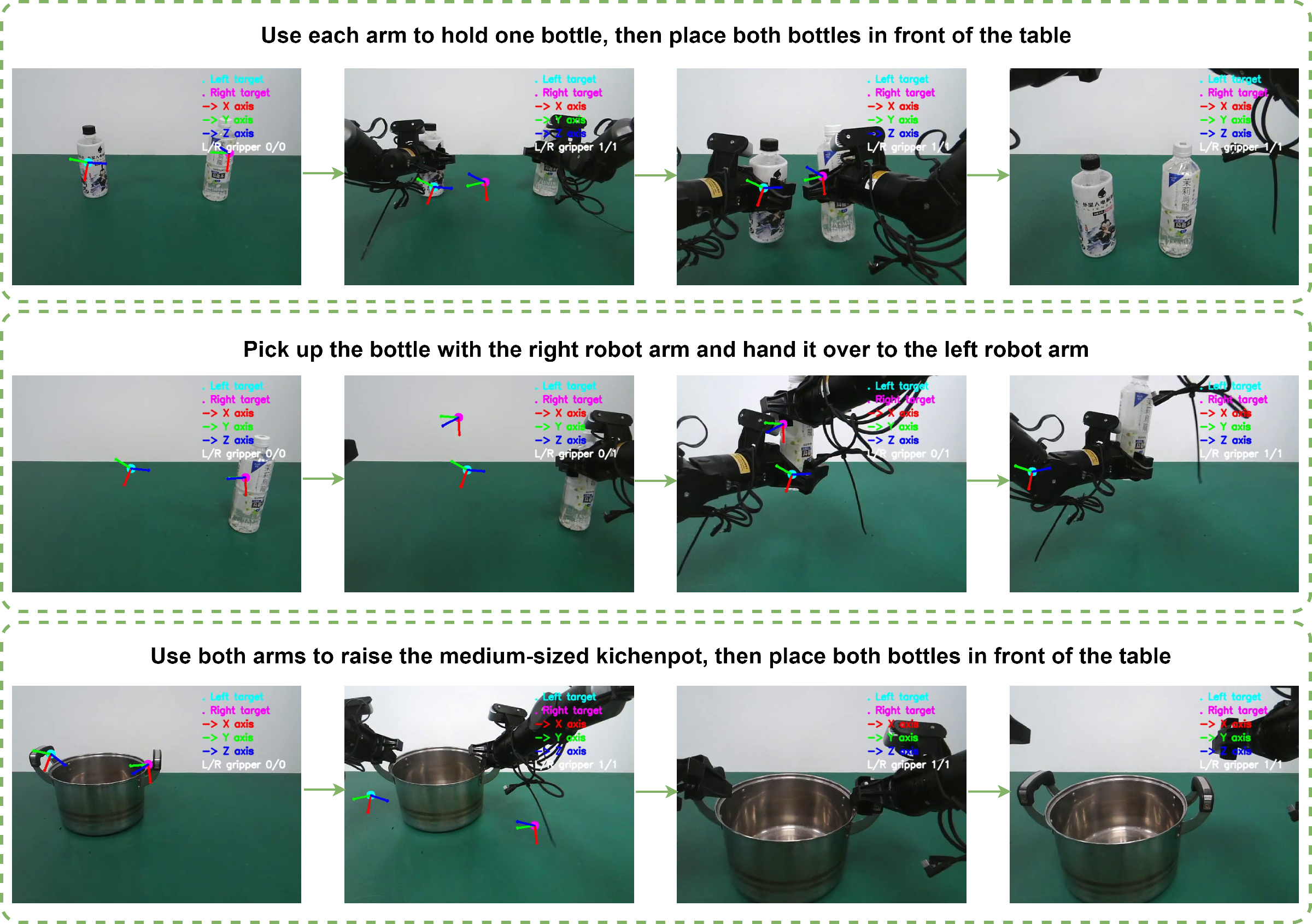}
    \caption{
    Key-action visualization on the Agilex robot. In the examples shown, the key-actions predicted for both the left and right arms are consistent with the intended manipulation stages.
    }
    \label{fig:real_key_vis}
\end{figure}

In Fig.~\ref{fig:real_key_vis}, we visualize the predicted key-actions during the real-world experiments across the three dual-arm tasks.
In the examples shown, the predicted key-actions correspond to intermediate targets for both arms across temporally distinct manipulation stages.
This is especially important for complex bimanual tasks such as handover, where sequential right-to-left inter-arm coordination is required.


\end{document}